\PassOptionsToPackage{unicode}{hyperref}
\PassOptionsToPackage{hyphens}{url}
\documentclass[14pt]{article}
\usepackage{amsmath,amssymb}
\usepackage{lmodern}
\usepackage{makecell}  % Add this in the preamble
\usepackage{iftex}
\ifPDFTeX
  \usepackage[T1]{fontenc}
  \usepackage[utf8]{inputenc}
  \usepackage{textcomp} % provide euro and other symbols
\else % if luatex or xetex
  \usepackage{unicode-math}
  \defaultfontfeatures{Scale=MatchLowercase}
  \defaultfontfeatures[\rmfamily]{Ligatures=TeX,Scale=1}
\fi
\IfFileExists{upquote.sty}{\usepackage{upquote}}{}
\IfFileExists{microtype.sty}{% use microtype if available
  \usepackage[]{microtype}
  \UseMicrotypeSet[protrusion]{basicmath} % disable protrusion for tt fonts
}{}
\makeatletter
\@ifundefined{KOMAClassName}{% if non-KOMA class
  \IfFileExists{parskip.sty}{%
    \usepackage{parskip}
  }{% else
    \setlength{\parindent}{0pt}
    \setlength{\parskip}{6pt plus 2pt minus 1pt}}
}{% if KOMA class
  \KOMAoptions{parskip=half}}
\makeatother
\usepackage{xcolor}
\usepackage{algorithm}
\usepackage{algpseudocode}
\usepackage{longtable,booktabs,array}
\usepackage{calc} % for calculating minipage widths
\usepackage{etoolbox}
\makeatletter
\patchcmd\longtable{\par}{\if@noskipsec\mbox{}\fi\par}{}{}
\makeatother
\IfFileExists{footnotehyper.sty}{\usepackage{footnotehyper}}{\usepackage{footnote}}
\makesavenoteenv{longtable}
\usepackage{graphicx}
\makeatletter
\def\maxwidth{\ifdim\Gin@nat@width>\linewidth\linewidth\else\Gin@nat@width\fi}
\def\maxheight{\ifdim\Gin@nat@height>\textheight\textheight\else\Gin@nat@height\fi}
\makeatother
\setkeys{Gin}{width=\maxwidth,height=\maxheight,keepaspectratio}
\makeatletter
\def\fps@figure{htbp}
\makeatother
\ifLuaTeX
  \usepackage{selnolig}  % disable illegal ligatures
\fi
\ifPDFTeX
  \newcommand{\RL}[1]{\beginR #1\endR}

\fi
\IfFileExists{bookmark.sty}{\usepackage{bookmark}}{\usepackage{hyperref}}
\IfFileExists{xurl.sty}{\usepackage{xurl}}{} % add URL line breaks if available
\hypersetup{
  hidelinks,
  pdfcreator={LaTeX via pandoc}}

\author{}
\date{}
\usepackage{geometry}
\usepackage{setspace}
\usepackage{titlesec}
\usepackage{lmodern}
\usepackage{graphicx}
\usepackage{hyperref}

\titleformat{\section}{\large\bfseries}{\thesection}{0.5 em}{}
\titleformat{\subsection}{\normalsize\bfseries}{\thesubsection}{0.5em}{}

\begin{document}

\begin{center}
    \Large \textbf{PINN-DT: Optimizing Energy Consumption in Smart Buildings Using Hybrid Physics-Informed Neural Networks and Digital Twin Framework with Blockchain Security} \\[0.9 em]
    
\normalsize

\textbf{Hajar Kazemi Naeini}\textsuperscript{1}, 
\textbf{Roya Shomali}\textsuperscript{2}, 
\textbf{Abolhassan Pishahang}\textsuperscript{3}, 
\textbf{Hamidreza Hasanzadeh}\textsuperscript{4},
\textbf{Mahdieh Mohammadi}\textsuperscript{5}, 
\textbf{Saeed Asadi}\textsuperscript{1}, 
\textbf{Abbas Varmaghani}\textsuperscript{6}, 
\textbf{Ahmad Gholizadeh Lonbar}\textsuperscript{7,*} \\[0.5em]

\footnotesize
    \textsuperscript{1}Department of Civil Engineering, University of Texas at Arlington, Arlington, Texas, USA\\
    \textsuperscript{2}Department of Information Systems, Statistics and Management Science, The University of Alabama, USA\\
    \textsuperscript{3}College of Arts and Letters, School of Art, Florida Atlantic University, Boca Raton, FL, USA\\
    \textsuperscript{4}Department of Environment and Energy, Science and Research Branch, Islamic Azad University, Tehran, Iran\\
    \textsuperscript{5}Department of Quantity Surveying, University of Malaya, Kuala Lumpur, Malaysia\\
    \textsuperscript{6} Department of Computer Engineering, Islamic Azad University of Hamadan, Hamedan, Iran
\\[0.3em]  
    \textsuperscript{7}Department of Civil, Construction, and Environmental Engineering, University of Alabama, Tuscaloosa, AL, USA\\[0.3em]
    *Corresponding author: \href{mailto:agholizadehlonbar@crimson.ua.edu}{agholizadehlonbar@crimson.ua.edu}
\end{center}

\vspace{0.5em}

\noindent \textbf{Abstract} \\
The advancement of smart grid technologies necessitates the integration of cutting-edge computational methods to enhance predictive energy optimization. This study proposes a multi-faceted approach by incorporating (1) Deep Reinforcement Learning (DRL) agents trained using data from Digital Twins (DTs) to optimize energy consumption in real time, (2) Physics-Informed Neural Networks (PINNs) to seamlessly embed physical laws within the optimization process, ensuring model accuracy and interpretability, and (3) Blockchain (BC) technology to facilitate secure and transparent communication across the smart grid infrastructure.The model was trained and validated using comprehensive datasets, including smart meter energy consumption data, renewable energy outputs, dynamic pricing, and user preferences collected from IoT devices. The proposed framework achieved superior predictive performance with a Mean Absolute Error (MAE) of 0.237 kWh, Root Mean Square Error (RMSE) of 0.298 kWh, and an R-squared ($R^2$) value of 0.978, indicating a 97.8\% explanation of data variance. Classification metrics further demonstrated the model’s robustness, achieving 97.7\% accuracy, 97.8\% precision, 97.6\% recall, and an F1 Score of 97.7\%. Comparative analysis with traditional models like Linear Regression, Random Forest, SVM, LSTM, and XGBoost revealed the superior accuracy and real-time adaptability of the proposed method. In addition to enhancing energy efficiency, the model reduced energy costs by 35\%, maintained a 96\% user comfort index, and increased renewable energy utilization to 40\%. This study demonstrates the transformative potential of integrating PINNs, DTs, and Blockchain technologies to optimize energy consumption in smart grids, paving the way for sustainable,and efficient energy management systems.


\section{1. Introduction}

Current energy system development, coupled with enhanced emphasis on
sustainability, underscores the necessity for novel strategies towards
the enhancement of energy efficiency in smart grids and buildings.
Buildings and smart grids are central elements in the mitigation of the
global energy crisis, a situation worsened by escalating greenhouse gas
emissions and mounting energy demands. In this regard, the integration
of Machine Learning (ML) and Digital Twin (DT) technologies shows much
potential for energy conservation, cost saving, and better environmental
sustainability. Household appliances (HAs), in particular,
energy-intensive appliances like washing machines (WMs) and air
conditioners (ACs), account for approximately 30\% of overall energy
consumption in the United States {[}1{]}. Effective management of energy
consumption by residential energy management (REM) systems is necessary
to save energy costs as well as to maintain grid stability. REM systems
are made more complex by integrating distributed energy resources (DERs)
like solar photovoltaic (PV) panels, electric vehicles (EVs), and energy
storage systems (ESS). The above developments call for the development
of sophisticated Home Energy Management Systems (HEMS) that can improve
energy consumption while being mindful of user preferences and comfort
levels. In general, HEMS rely on two fundamental aspects: monitoring
energy consumption through smart meters and scheduling energy usage of
individual appliances in an optimized way. Traditionally, these systems
have been implemented using deterministic optimization methods, such as
mixed-integer nonlinear programming (MINLP) and mixed-integer linear
programming (MILP) {[}2-4{]}. While effective, these methods are limited
by their high computational complexity and challenges associated with
managing uncertainties in both user behavior and energy supply. The fast
development of data-driven technologies, such as Machine Learning (ML)
and Artificial Intelligence (AI), has introduced new opportunities for
the advancement of Renewable Energy Management (REM) systems.
Reinforcement Learning (RL), a branch of ML, has become an effective
approach to optimizing energy use in smart buildings. Google DeepMind
showed the promise of reinforcement learning (RL) in slashing data
center energy expenses by 40\% through innovative energy management
techniques {[}5{]}. 
Jamali and Abbasalizadeh [78] developed strategies for placing services within IoT systems while considering cost and performance trade-offs through multicriteria decision-making methods. Their results showed that strategic service co-location can substantially cut operational expenses without compromising efficiency, emphasizing smarter resource deployment in large-scale IoT networks.
Abbasalizadeh et al. [79] explored the integration of fuzzy logic with deep learning for dynamic scheduling in wireless communication systems. Their hybrid AI approach demonstrated improved adaptability and latency reduction under heavy network loads, suggesting a promising avenue for enhancing wireless network performance. Furthermore, techniques like Deep Q-Networks (DQN)
and policy gradient techniques have been used to improve building energy
efficiency {[}6, 7{]}. Although promising, current techniques frequently
overlook appliances\textquotesingle{} and distributed energy
resources\textquotesingle{} (DERs\textquotesingle) constant and diverse
operation, along with user comfort. Moreover, the rising penetration of
Renewable Energy Sources (RESs) and the growing system complexity have
necessitated the demand for more scalable and flexible solutions. It is
here that Digital Twin (DT) technology, in conjunction with Machine
Learning (ML), has the potential to effect revolutionary change.

Digital Twin (DT) technology, first defined by Grieves in 2002, provides
a virtual representation of physical systems for real-time monitoring,
evaluation, and control {[}1{]}. DT systems leverage data from sensors,
Internet of Things (IoT) devices, and advanced computational models to
create dynamic, virtual representations of physical assets. In the
energy industry, DT technology promises significant potential to address
issues related to optimization, reliability, and sustainability [74].

The integration of DT systems in smart grids enables high-level functions
such as fault detection, load forecasting, operator behavior, and health
monitoring of the energy system {[}3{]}. DTs also assist in real-time
decision-making through the bridge established between physical and
digital twins. This role is particularly crucial in managing complex
systems such as microgrids, transport systems, and distributed energy
systems.

Within the context of transportation infrastructures, digital twins
(DTs) have the capability to improve energy systems by providing timely
data on electric vehicle charging stations, traffic flow, and power
needs {[}4{]}. In microgrids, DTs ensure remote monitoring, prediction
maintenance, and efficient electricity distribution, thus strengthening
system resilience and reliability. The incorporation of Machine Learning
(ML) technologies in DT systems enhances their capabilities through
enhanced data analytics, forecasting, and data-driven decisions. In ML,
diverse algorithmic models, including their application in neural
networks, reinforcement learning, and deep learning, have capabilities
to handle massive amounts of both current and historical data,
ultimately to optimize power consumption, predict power demands, and
improve system efficiencies. An excellent case in point is application
in DT systems through the use of reinforcement learning (RL) to optimize
power use.Figure 1 shows the SDN-based Digital Twin architecture for smart energy systems. RL is designed to optimize power use in response to
constantly changing dynamics in the power industry, including power
price volatility, variability in renewable power generation, and shifts
in power-user behavior. Using insights derived through current and
historical data, RL-powered DT systems can develop optimal power use
strategies to optimize cost savings, efficiencies, and power-user
satisfaction. In addition, deep learning (DL) technologies, in the form
of convolutional neural networks (CNNs),  physics-informed neural networks (PINNs)[80] and recurrent neural networks
(RNNs), have found application in power forecasting and power faults in
smart grids {[}10-12{]}.These technologies allow for accurate power needs forecasting and power
supplies, thus enabling forward planning for power management. 

In spite
of their potential to optimize power systems through application,
several challenges have remained. The successful implementation of DT
systems demands seamless data fusion of data gathered through various
means, including sensors, IoT systems, and historical data. Norcéide et al. [75] examined how neuromorphic vision sensors enhance object tracking in augmented reality contexts. Their study demonstrated that event-driven vision technologies deliver superior tracking precision and energy efficiency relative to conventional camera-based methods, showcasing the value of these sensors for real-time AR applications. The data
collection and data-processing steps in such contexts bring serious
challenges. This study explores the potential of The Hybrid PINNs-DT
framework aims to address the limitations of existing deterministic and
ML-based methods by incorporating physical laws into the learning
process. This fusion enables better handling of uncertainties in user
behavior, renewable energy availability, and dynamic grid conditions
while maintaining computational efficiency.

\begin{figure}[htbp]
\centering
\includegraphics[width=4.8in,height=6.45in]{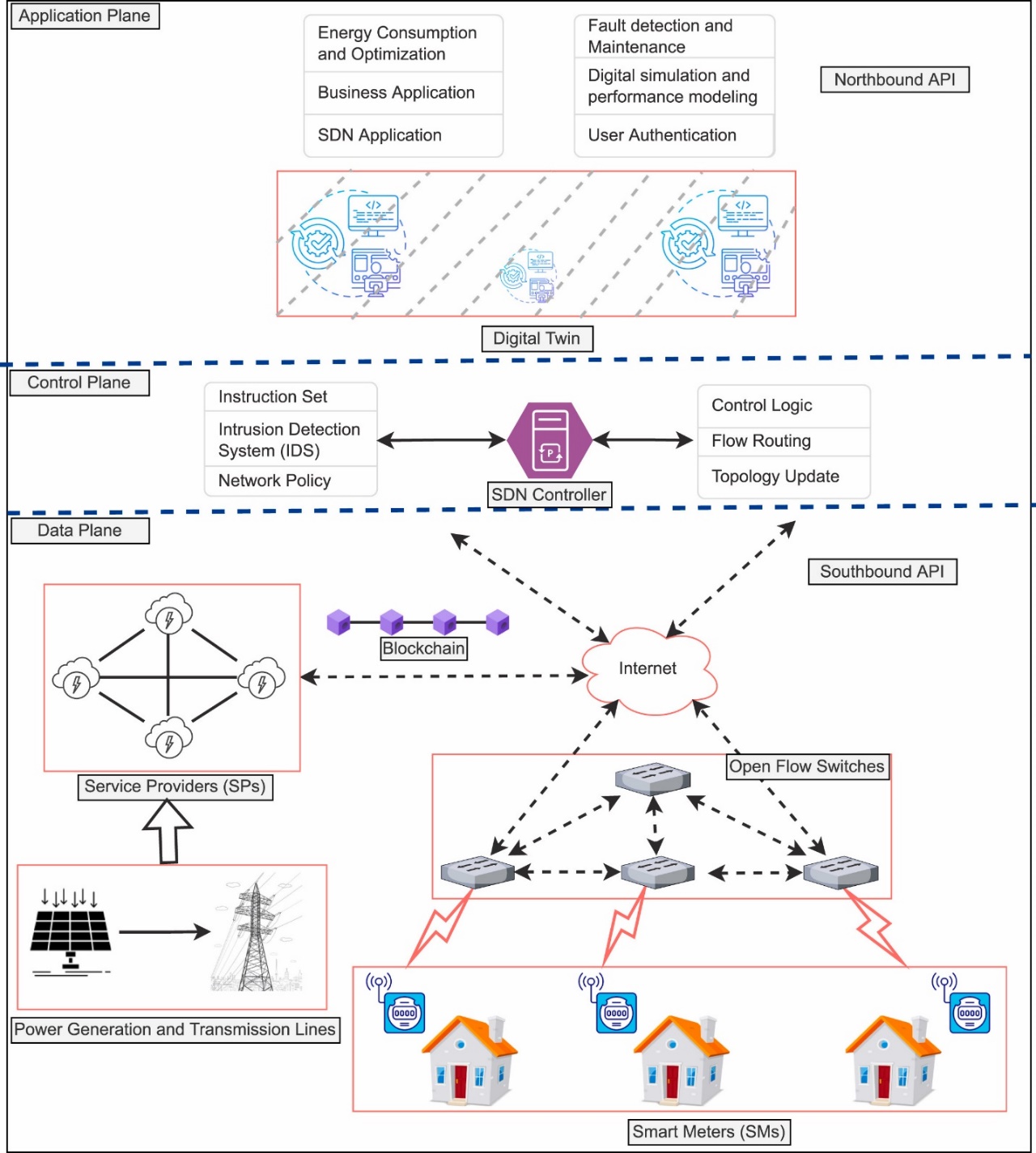}
\caption{Architecture of a Software-Defined Networking (SDN)-Based Digital Twin for Smart Energy Systems.}
\label{fig:sdn_dt_architecture}
\end{figure}

 The specific objectives of
this study are as follows:

\begin{enumerate}
\def\labelenumi{\arabic{enumi}.}
\item
  To review the current state of research on integrating PINNs-DT
  technologies into energy systems.
\item
  To identify the challenges and opportunities associated with
  implementing PINNs-DT systems in smart grids, particularly in the
  context of secure, real-time data exchange and scalable energy
  optimization.
\item
  To propose and validate innovative methods for combining PINNs-DT, and
  Blockchain technologies to enhance energy efficiency, reliability, and
  sustainability in smart grids.
\end{enumerate}

These objectives, the proposed method aims to contribute to the
development of intelligent energy management systems that balance
economic efficiency with user comfort, enhance cybersecurity, and
support the transition toward sustainable and carbon-neutral energy
infrastructures.

\section{2. Related work}

A thorough evaluation of real-time analytic techniques in digital twins
was given by Haghi et al. {[}14{]}, who focused on physics-informed
modeling, data-driven simulations, and machine learning applications to
speed up and minimize delays in digital twin calculations. The function
of digital twins in optical networks was studied by Wang et al.
{[}15{]}, who described their architecture for automated control, mirror
modeling, and real-time monitoring. Future research directions and
developments in intelligent network automation are highlighted in this
paper. A microgrid digital twin framework that integrates IoT, AI, and
big data analytics was presented by Utama et al. {[}16{]} and is based
on the Smart Grid Architectural Model (SGAM). Their case study showed
enhanced energy management effectiveness and interoperability.

In their discussion of digital twin applications in the wind energy
sector, Stadtmann et al. {[}17{]} identified important industry issues
such regulatory requirements, modeling limitations, and data
dependability. They put up a plan for upcoming developments and industry
adoption. For hydropower management, Zeng et al. {[}18{]} suggested a
hybrid system that combines neural networks, digital twins, and type-2
fuzzy logic controllers. Their approach reduced maintenance costs,
increased operating efficiency, and enhanced defect detection. In their
assessment of AI-powered civil engineering applications, Xu et al.
{[}19{]} highlighted the application of AI in smart city management,
structural health monitoring, and design optimization. They tackled
integration issues including data security and scalability. Ahmadi et
al. {[}20{]} integrated Finite Element Analysis (FEA) with
Physics-Informed Neural Networks (PINNs) to enhance the biomechanical
modeling of the human lumbar spine. Their approach automates spine
segmentation and meshing, addressing challenges in material property
prediction. The development of cyber-physical power systems was examined
by Parizad et al. {[}21{]}, who described how AI, blockchain, and IoT
are integrated into contemporary power networks. They emphasized the
difficulties in maintaining control, security, and stability in the
delivery of energy.

Attari et al. {[}44{]} proposed an advanced optimization framework
employing mathematical modeling and meta-heuristic algorithms to
optimize inventory logistics in reverse warehouse systems, focusing on
reducing costs and enhancing storage efficiency. In alignment with
sustainable development goals, Asadi et al. {[}45{]} reviewed analytical
and numerical approaches in earth-to-air heat exchangers, categorizing
methods into analytical, numerical, and exergeoeconomic areas to enhance
thermal efficiency and reduce operational costs. Moghim \& Takallou
{[}46{]} assessed extreme hydrometeorological events in Bangladesh using
the Weather Research and Forecasting model. Their study identified the
efficiency of Bayesian regression in improving rainfall predictions,
enhancing early warning systems. Complementing these sustainability
strategies, Asgari et al. {[}47{]} explored the critical relationship
between energy consumption and GDP through threshold regression
analyses, underlining the importance of energy-efficient growth and
sustainable development strategies.

\begin{table}[htbp]
\centering
\caption{ Summary of Literature Review on PINN and Related Applications}
\small
\begin{tabular}{p{1cm}lp{2.5cm}p{3cm}p{4cm}}
\toprule
\textbf{Author} & \textbf{Year} & \textbf{Method} & \textbf{Aim} & \textbf{Result} \\
\midrule
Chen et al. [23] & 2025 & Physics-informed encoder-decoder model & Predict carbon emissions and identify anomalies & Improved accuracy by 9.24\% with enhanced robustness \\
Chen et al. [24] & 2025 & AI applications in sustainable energy & Review AI's role in multi-energy systems & Identified challenges and proposed layered security strategies \\
Mittal et al. [25] & 2025 & Physics-informed neural network & Detect and classify wild animal activity & Achieved high accuracy and real-time alert generation \\
Pandiyan et al. [26] & 2025 & Physics-informed neural network (PINN) & Optimize electric water heater modeling & Enhanced computational efficiency and performance \\
Feng et al. [27] & 2025 & Uniform Physics-Informed Neural Network (UPINN) & Extract parameters for voltage stability & Improved accuracy in real-time voltage stability monitoring \\
Habib et al. [28] & 2025 & Block-based physics-informed neural network & Estimate inelastic response of base-isolated structures & Reduced computational cost and improved predictive performance \\
Nadal et al. [29] & 2025 & Physics-Informed Neural Networks (PINNs) & Enhance simulation accuracy in power system dynamics & Improved predictive precision in power system simulations \\
Ventura Nadal et al. [30] & 2025 & Physics-Informed Neural Networks (PINNs) & Improve power system simulation accuracy & Enhanced modeling and reduced computational error \\
Ko et al. [31] & 2025 & Physics-Informed Neural Networks (PINNs) & Long-term prognostics of proton exchange membrane fuel cells & Achieved high accuracy in fuel cell lifespan prediction \\
Qin et al. [32] & 2024 & Inverse Physics-Informed Neural Networks (PINNs) & Develop a digital twin-based approach for bearing fault diagnosis under imbalanced samples & Enhanced fault diagnosis accuracy and improved precision in cross-working-condition detection \\
\bottomrule
\end{tabular}
\end{table}

\section{3. The Concept of Digital Twin (DT)}

\subsection{\textbf{3.1 Introduction to Digital Twin Technology}}

Digital Twin (DT) technology has emerged as a groundbreaking innovation
bridging the physical and digital realms. The concept, first introduced
in 2002 by Grieves for product lifecycle management {[}32{]}, provides a
dynamic digital representation of physical entities, systems, or
processes. This digital replica enables real-time monitoring, analysis,
and optimization, offering insights into behaviors and dynamics that
were previously unattainable {[}33{]}. By creating a virtual counterpart
of a physical system, DTs serve as a powerful tool for predictive
maintenance, fault detection, optimization, and simulation,
revolutionizing industries such as energy, manufacturing, healthcare,
and transportation. As energy systems become more complex, ensuring the
scalability, interoperability, and security of Digital Twin (DT) systems
is critical. This includes the integration of DT systems with energy
management platforms and the accommodation of diverse user
needs.{[}34{]}. The extensive use of data in DT systems highlights the
importance of addressing cybersecurity and data privacy concerns.
Ensuring secure and efficient data exchange, particularly through
blockchain technology, and protecting sensitive information are
paramount to the success of such systems {[}35{]}. Figure 2 illustrates
the proposed multi-layered architecture that integrates Software-Defined
Networking (SDN), DT technology, Deep Reinforcement Learning (DRL), and
Blockchain into smart energy systems. The segmentation models highlights the importance of selecting architecture-specific solutions within digital twin environments, where accurate and real-time anatomical modeling is essential for clinical decision support [77]. The architecture consists of three
planes, where the Application Plane hosts energy optimization, fault
detection, digital simulation, and user authentication processes,
interfacing with lower layers via the Northbound API\textbf{.} The
integration of advanced machine learning (ML) techniques and edge
computing with DT systems addresses challenges related to scalability,
computational efficiency, and real-time decision-making. By leveraging
the predictive and analytical capabilities of ML and the secure
framework provided by Blockchain, the proposed DT system enables
proactive energy optimization, real-time fault detection, and efficient
energy distribution while ensuring robust cybersecurity (see Table 1).
The integration of Large Language Models (LLMs) into Digital Twin systems offers a promising avenue for enhancing contextual understanding, user interaction, and decision-making across smart grid and energy management applications.Farhadi Nia et al. [74] studied the integration of ChatGPT and LLMs in dental diagnostics. Their findings show that LLMs enhance clinical decision-making, streamline patient-provider communication, and improve procedural efficiency in oral healthcare.

\begin{figure}[htbp]
\centering
\includegraphics[width=4.43in,height=3.88in]{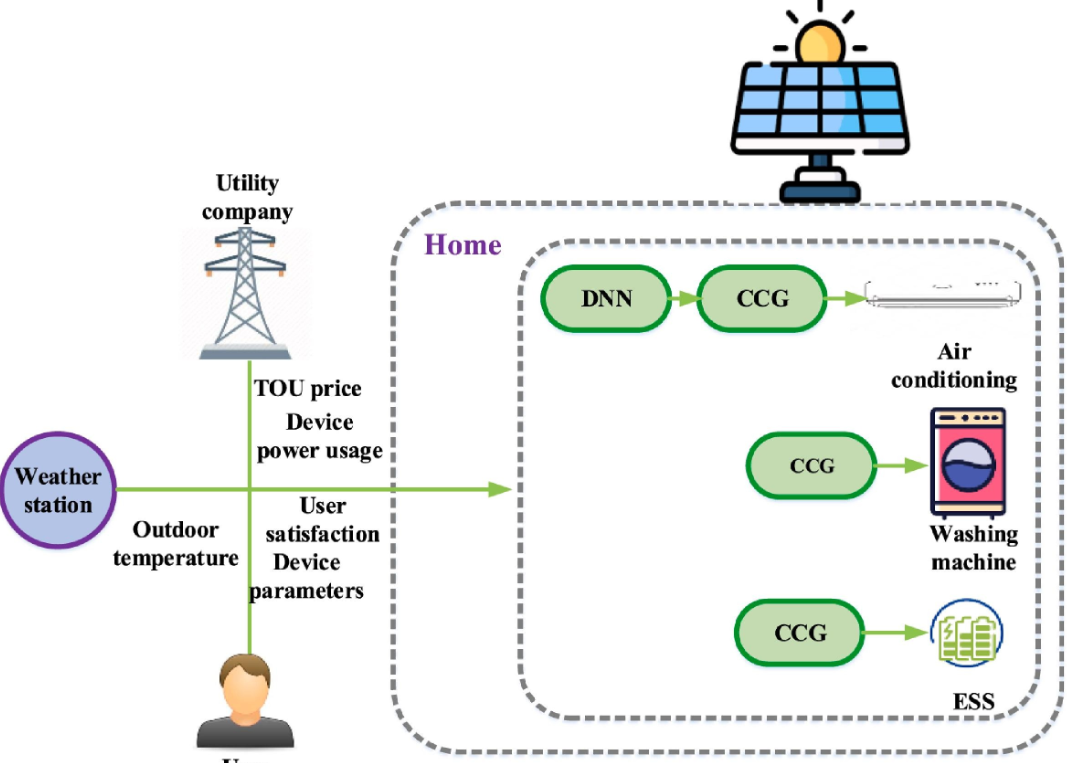}
\caption{A smart home energy system using DNN and CCG to optimize appliances with real-time data}
\label{fig:smart_home_dnn_ccg}
\end{figure}

DT technology lies in its ability to provide actionable insights by
integrating data, analytics, and simulation capabilities. By leveraging
real-time data streams, DT systems can anticipate potential issues,
optimize operations, and improve overall system performance. This
ability makes DT technology a cornerstone of digital transformation
across various sectors. Advanced ML algorithms and DT models often
require significant computational resources.

\subsection{\textbf{3.2 Key Components of Digital Twin Technology}}

The DT prototype serves as the foundational digital representation of a
physical entity. It includes all essential virtual data, such as
properties, designs, parameters, and configurations, necessary for
creating an accurate and functional digital model. The prototype acts as
a blueprint for developing DT instances, ensuring consistency and
accuracy in representing physical systems {[}36{]}.
Optimize codec efficiency can be adapted within digital twin architectures to enhance real-time data processing and reduce bandwidth consumption in complex simulation environments [76]. DT instances are
specific digital models linked to their physical counterparts throughout
their lifecycle. These instances are updated continuously with real-time
data, reflecting the current state of the physical system. By
maintaining synchronization, DT instances enable real-time monitoring,
predictive analysis, and decision-making for individual assets {[}37{]}.
The DT aggregate enchases all individual DT instances and prototypes,
creating a unified representation of complex systems. Aggregates allow
for holistic analysis and simulation of interconnected components,
enabling a comprehensive understanding of system behaviors and
interactions {[}38{]}. The DT environment cotises the hardware,
software, and network infrastructure required to support DT systems.
This includes IoT devices, sensors, simulation tools, and data analytics
platforms. The environment facilitates real-time data collection,
processing, and visualization, ensuring seamless interactions between
the physical and digital realms {[}38{]}.

\subsection{\textbf{3.3 Core Functions of Digital Twin Technology}}

Data integration is the backbone of DT technology. Sensors, gauges, RFID
tags, cameras, and other devices collect data from physical systems,
which is then transmitted to the DT system in real-time or with minimal
delay. This comprehensive data integration ensures accurate and reliable
digital representations. Advanced simulation tools model the behaviors
and interactions of physical systems under various conditions. This
enables predictive analysis, optimization, and scenario planning,
providing valuable insights for decision-making {[}39{]}. By leveraging
AI and ML algorithms, DT systems offer powerful analytics capabilities.
These include predictive maintenance, anomaly detection, and
optimization strategies, which improve system reliability and
performance. Visualization tools provide user-friendly interfaces to
interpret complex data and simulation results. These tools enable
stakeholders to analyze system behaviors, identify trends, and make
informed decisions effectively {[}40{]}.

\subsection{\textbf{3.4 Evolution of Digital Twin Technology}}

The concept of DT was first formalized in a roadmap published by NASA in
2010 for health management of flight systems {[}19{]}. Early
applications focused on improving reliability and performance through
simulation and data integration. Grieves introduced the
three-dimensional DT model, consisting of the physical entity, its
virtual representation, and the data connections between them. This
model emphasized real-time synchronization and data-driven
decision-making. Tao and Zhang expanded the DT model to include five
dimensions: physical entity, virtual model, services, fusion data, and
their interconnections {[}41{]}. This enhanced model supported
cross-domain integration and reusability, enabling diverse industrial
applications. The integration of IoT, AI, and cyber-physical systems has
significantly advanced DT technology. These technologies enable
real-time data collection, advanced analytics, and seamless
interactions, enhancing the capabilities and applications of DT systems
{[}42{]}.

\subsection{\textbf{3.5 Applications for Digital Twin Technology}}

DT technology is pivotal in the renewable energy sector, where it aids
in fault detection, performance optimization, and predictive
maintenance. For example, digital replicas of solar PV cells can detect
defects caused by cell degradation or mismatched modules, improving
system efficiency and reliability {[}43{]}. In smart grids, DTs enhance
system reliability by enabling real-time monitoring, predictive
analytics, and optimization. DT systems are applied at unit, system, and
system-of-systems (SoS) levels to optimize processes such as power
generation, transmission, distribution, and consumption {[}44{]}. DTs
facilitate efficient energy management in transportation networks,
particularly in electric vehicle (EV) charging infrastructure. By
integrating real-time data from traffic patterns and charging stations,
DT systems optimize energy distribution and support sustainable
transportation solutions {[}45{]}. In manufacturing, DT technology
supports product design, production planning, and equipment maintenance
{[}47{]}. By simulating production processes, DTs enable predictive
maintenance and operational optimization, reducing downtime and costs
{[}48-50{]}. DTs are increasingly used in healthcare to create virtual
models of human organs and systems. These models support personalized
treatment plans, surgical simulations, and disease monitoring, enhancing
patient outcomes {[}51-54{]}.

\begin{table}[htbp]
\centering
\caption{Summary of Literature Review on Blockchain, Digital Twin, and Energy Systems}
\small
\begin{tabular}{p{1cm}p{2.3cm}p{1.5cm}p{3cm}p{3cm}}
\toprule
\textbf{Authors} & \textbf{Methodology} & \textbf{Platform} & \textbf{Outcome} & \textbf{Challenge} \\
\midrule
Zahid et al. [56] & AI, Digital Twins, Blockchain, Metaverse & Smart Grid 3.0 & Enhanced real-time monitoring, decentralized transactions, and system automation & Interoperability, scalability, cybersecurity, and data integrity issues \\
Sarker et al. [57] & Explainable AI (XAI) and cybersecurity modeling & Digital Twin Environments & Improved AI-driven cybersecurity automation and threat detection & Ensuring trustworthiness, human-explainability, and AI transparency \\
Idrisov et al. [58] & ML and Digital Twin-based anomaly detection & Power Electronics Dominated Grids (PEDGs) & Real-time tracking of grid anomalies and cyberattack prevention & Handling complex grid operations and cybersecurity vulnerabilities \\
Meng et al. [59] & IoT, Blockchain, Cybersecurity & Smart Urban Energy Systems & Enhanced cybersecurity and efficient energy management in urban grids & Integration complexity and real-time cyber threat mitigation \\
Kavousi-Fard et al. [60] & Digital Twin for Renewable Energy Resources (RER) & Solar Energy Systems & Optimized energy management and real-time monitoring of solar grids & Variability in energy generation and reliability challenges \\
Kabir et al. [61] & IoT-Driven Digital Twin Systems & Smart Energy Grids & Improved operational efficiency, predictive maintenance, and grid sustainability & Infrastructure compatibility and data security concerns \\
Cali et al. [62] & Cybersecurity, Digital Twins, AI & Energy Systems and Smart Cities & Enhanced efficiency, security, and sustainability in energy infrastructure & Ensuring secure real-time data transmission and system resilience \\
Jafari et al. [63] & Multi-Layer Digital Twin Model & Smart Grid, Transportation, Smart Cities & Reliable energy distribution and improved grid operations & Managing real-time data flow and system scalability \\
\bottomrule
\end{tabular}
\end{table}

\subsection{\textbf{3.6 Traditional Grid}}

The traditional electrical grid operates as a centralized power
generation network that interconnects transmission and distribution
systems using electromechanical infrastructure {[}55-57{]}. This grid
delivers electricity over extensive areas through a one-way
transmission-distribution system controlled centrally using electrically
operated mechanical devices {[}63-67{]}. The centralized energy
infrastructure, with limited sensors, faces significant challenges in
monitoring, control, and self-healing capabilities. Manual monitoring
makes power distribution and transmission inefficient, leading to high
losses, difficulty in fault detection, prolonged outages, and economic
losses due to extended restoration times and grid overheating incidents
{[}57,60{]}.

\subsection{\textbf{3.7 Microgrid}}

A microgrid, an emerging technology, leverages Distributed Energy
Resources (DERs) to address the shortcomings of traditional electric
grids. By utilizing DERs, power transmission and distribution losses are
minimized, creating a more efficient, secure, and cost-effective energy
system. DERs enable the integration of renewable energy sources such as
solar, wind, and wave power, reducing reliance on coal and natural gas,
thereby supporting clean energy initiatives {[}36{]}. The microgrid acts
as a controlled segment of the grid, simplifying the complexities
associated with DERs and providing structured expansion opportunities to
enhance the grid's quality, security, and efficiency {[}22{]}. It
integrates distributed power grids systematically, optimizing operations
via the Point of Common Coupling (PCC) to ensure a reliable power system
{[}25,28,29{]}.Table 2 summarizes recent literature on the integration of Blockchain, Digital Twin, and energy systems.

A microgrid is defined as a localized collection of energy sources and
loads, operating either in conjunction with the main grid or
independently. In its grid-connected mode, it offers ancillary services
and ensures uninterrupted power supply by managing transitions between
connected and standalone modes. An isolated or ``standalone microgrid''
functions entirely independently of larger electrical networks {[}31{]}.
In its dual-mode capability, the microgrid can seamlessly switch between
grid-connected and autonomous modes. During power deterioration or
network contingencies, it connects or disconnects from the main grid
using the PCC network, delivering standard power services. It
continuously monitors small-scale generators, associated loads, energy
storage, sensors, measurement units, and control systems, forming a
unified controllable entity. DERs operate in two modes: grid-connected
and autonomous (islanded), with the latter serving as a transitional
state between these modes {[}33{]}. Microgrids may be constructed in AC,
DC, or hybrid configurations, offering features such as ``plug and
play'' and ``peer-to-peer'' functionality. While supporting renewable
energy sources, not all microgrids fully utilize these resources.
Protective devices such as reclosers, circuit breakers, and relays
manage fault isolation in traditional grids. In microgrids, leakage
current variations during mode transitions necessitate advanced
safeguarding mechanisms for Distributed Generation (DG) plants
{[}35,36{]}.

\subsection{\textbf{3.8 Smart Grid}}

The smart grid integrates communication, data storage, and analysis
capabilities to enable rapid, intuitive, and collaborative energy
network operations. Unlike traditional grids that rely on centralized
electricity generation and one-way power flow with high transmission
losses, smart grids utilize two-way information and power flows,
combining centralized and distributed systems. These advancements
enhance efficiency, reliability, and sustainability {[}68{]}. Smart
grids leverage modern communication and information technology (IT),
incorporating sensors, remote monitoring systems, control devices, and
domestic appliances connected to the grid. Technologies such as
Supervisory Control and Data Acquisition (SCADA) and synchrophasors
generate extensive data, requiring robust systems for handling,
analysis, and actionable insights {[}69,38,45{]}. Intelligent
electricity generation in smart grids employs advanced IT solutions to
improve energy efficiency, reliability, and security while supporting
renewable energy adoption and environmental goals. The figure 3
illustrates the integration of reality and a Digital Twin system for
energy management in various power infrastructures. The Reality layer
includes components such as substations, single-family detached,
multi-family residential buildings, open-space PV installations, and
wind farms. These physical entities are interconnected through a grid
network.

These systems interact with control hubs and energy supply structures to
monitor and analyze the power system in real time, reducing delays and
optimizing operations {[}45{]}. The smart grid's self-awareness,
self-optimization, and self-customization capabilities enable its
components to function autonomously or with minimal human intervention.
Instantaneous communication among systems, employees, and consumers
fosters a highly adaptive electricity generation model that
significantly enhances energy efficiency in the electrical sector.
\begin{figure}[htbp]
\centering
\includegraphics[width=4.9in,height=3.08in]{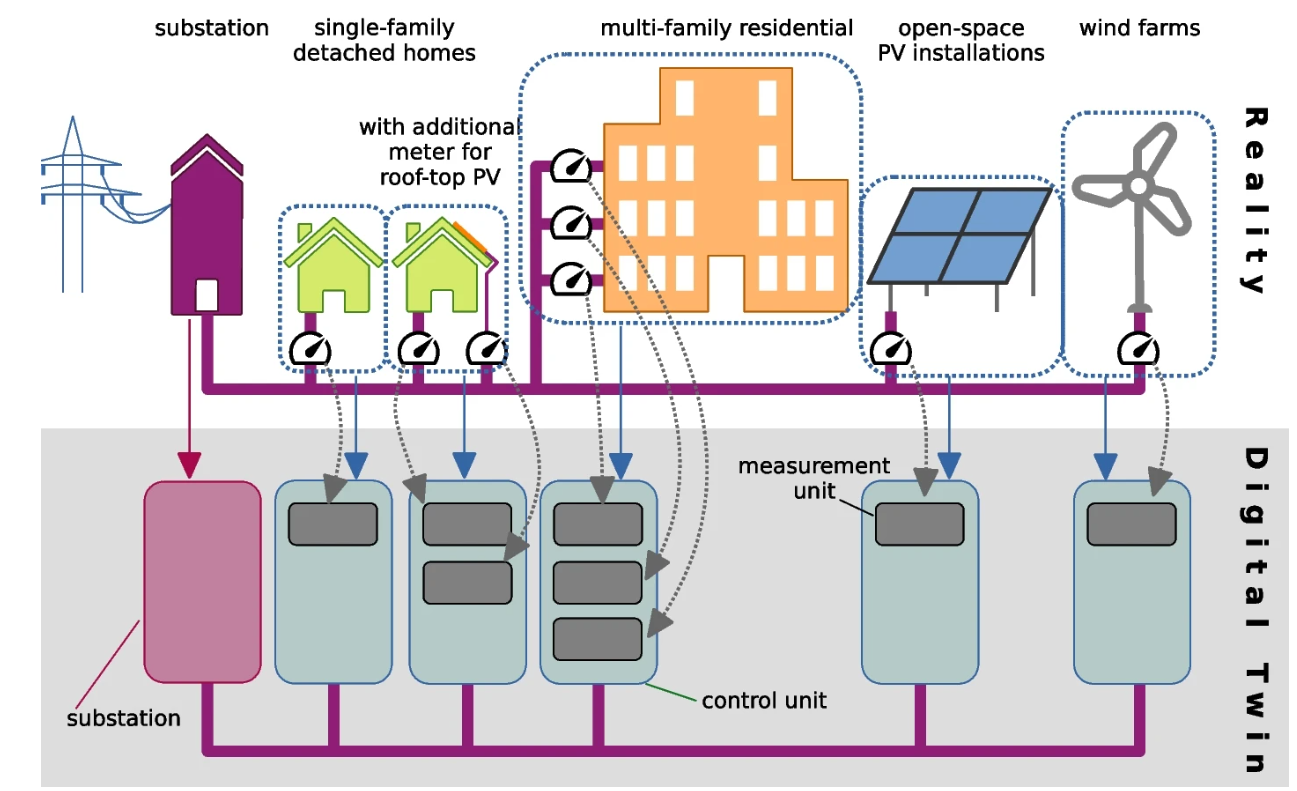}
\caption{The digital twin represents the real energy system's abstraction. Turquoise boxes denote control units, gray boxes represent measurement units, and the right side illustrates local wind farms and photovoltaic installations. Grey arrows show smart meters as measurement units, blue arrows map real entities to control units, and crimson arrows represent substations.}
\label{fig:dt_energy_abstraction}
\end{figure}

Despite its advantages, the transition from conventional to smart grids
involves high costs, posing challenges for industrial expenses {[}70{]}.
Additionally, cybersecurity risks, including potential data theft and
malicious attacks, remain a concern for smart grids utilizing
internet-based real-time information exchange {[}71,72{]}.

\section{4. Method and Materials}

A Digital Twin (DT) and Deep Reinforcement Learning (DRL) method is
proposed to optimize energy consumption in smart buildings while
maintaining occupant comfort and grid reliability. Multi-agent systems
are used to make real-time decisions about energy-intensive appliances.
As it interacts with the simulated grid, the DRL agent learns optimal
energy strategies. A blockchain-based decentralized data-sharing
mechanism ensures secure, real-time communication between devices, grid
components, and the DT system. Smart contracts are used to protect data
integrity and control access to it to address cybersecurity concerns.

An innovative method optimizes energy consumption in smart buildings by
combining Digital Twin (DT) technology and Deep Reinforcement Learning
(DRL). DT provides a high-fidelity virtual model of the building that
simulates its energy consumption patterns and integrates real-time data
from the IoT. Through interactions with this environment, a DRL agent
learns and executes optimal energy management policies, balancing cost
and user comfort. DT system and physical components communicate securely
and decentralized through a blockchain-based data sharing system. In
addition to enhancing energy efficiency and grid stability, the proposed
framework offers a scalable solution for future smart grid applications.

\subsection{\textbf{4.1. Dataset}}

To training and validating the machine learning-driven digital twin (DT)
system for energy optimization in smart buildings, a comprehensive and
multifaceted dataset is employed. The dataset is primarily based on data
collected from smart meters regarding detailed energy consumption
patterns. Smart meters provide granular-level information, including
timestamps, appliance identifiers, and energy consumption values (in
kWh), which are crucial for training the deep reinforcement learning
(DRL) agent to predict and optimize the scheduling of energy-intensive
appliances such as washing machines, air conditioners, and electric
heaters. \RL{}In addition to energy consumption data, the dataset
incorporates real-time data from solar photovoltaic (PV) panels and wind
turbines. This includes weather-related variables such as temperature,
solar irradiance, humidity, and wind speed, as well as the corresponding
energy outputs from these renewable sources. By leveraging this data,
the digital twin can accurately model the inherent uncertainties in
renewable energy generation, which is critical for optimizing the
integration and utilization of renewable energy sources in smart
buildings.

To further enhance the model\textquotesingle s adaptability, data from
smart IoT devices, including smart thermostats, occupancy sensors, and
lighting control systems, are integrated into the dataset. These devices
provide valuable insights into user preferences and comfort levels,
capturing variables such as preferred temperature settings, lighting
intensity, and appliance usage patterns. This user-centric data is
incorporated into the optimization equations as constraints or penalties
to ensure that energy efficiency measures do not compromise occupant
comfort. \RL{}The dataset also includes dynamic electricity pricing
information, grid demand, and supply fluctuations obtained from publicly
available sources. This real-time pricing data allows the DRL agent to
dynamically adjust energy consumption schedules to minimize costs,
particularly during peak demand periods or when energy prices fluctuate
significantly. Additionally, data on distributed energy resources
(DERs), such as electric vehicle (EV) charging patterns and energy
storage system (ESS) performance, are included to further refine the
energy management strategies. \RL{}To address the cybersecurity aspect
of the proposed framework, datasets like N-BaIoT are utilized. This
dataset includes network activity logs, timestamps, attack types, and
security labels that help in training the blockchain-enabled security
framework to detect and mitigate cyber threats within smart grid
networks. The integration of blockchain technology ensures secure,
transparent, and tamper-proof communication between all stakeholders
involved in the energy system. \RL{}the combined dataset captures a wide
range of variables, including building energy systems, user behavior,
renewable energy generation, dynamic grid conditions, and cybersecurity
metrics. This holistic approach ensures that the proposed framework can
effectively model and optimize complex interactions between these
factors. As a result, the system can achieve enhanced energy efficiency,
cost reduction, user satisfaction, and robust cybersecurity while
maintaining scalability and reliability in smart grid applications.

\subsection{\textbf{4.2. Hybrid Physics-Informed Neural Networks (PINNs) and
Digital Twin (DT) for Energy Optimization}}

To further enhance energy optimization in smart buildings and grids, we
propose integrating Hybrid Physics-Informed Neural Networks (PINNs) with
Digital Twin (DT) technology. This approach leverages the strengths of
physics-based modeling and data-driven techniques to achieve more
accurate, efficient, and adaptive energy management.

\subsection{\textbf{4.2.1. Hybrid PINNs-DT Framework}}

The Hybrid PINNs-DT framework aims to address the limitations of
existing deterministic and ML-based methods by incorporating physical
laws into the learning process. This fusion enables better handling of
uncertainties in user behavior, renewable energy availability, and
dynamic grid conditions while maintaining computational efficiency.

\begin{itemize}
\item
  Physics-Informed Neural Networks (PINNs) incorporate governing
  physical equations, such as thermodynamics, fluid dynamics, and
  electrical circuit laws, directly into the neural network's loss
  function. This ensures that the model adheres to known physical
  principles while learning from data, resulting in more accurate and
  generalizable predictions.
\item
  Digital Twin (DT) provides a real-time virtual replica of the physical
  energy system, integrating data from IoT sensors, smart meters, and
  DERs. It continuously updates the state of the system, allowing for
  dynamic simulation, monitoring, and optimization.
\item
  Reinforcement Learning (RL) algorithms, such as Deep Q-Networks (DQN)
  and Policy Gradient Methods, are integrated into the framework to
  optimize decision-making processes. The RL agent interacts with the DT
  environment, learning optimal energy management strategies over time.
\item
  Blockchain Integration ensure secure and transparent data exchange,
  blockchain technology is incorporated. This decentralized approach
  safeguards data integrity and supports trust among various
  stakeholders, including energy providers, consumers, and regulatory
  bodies.
\end{itemize}

\subsection{\textbf{4.2.2. Methodology}}

The optimization objective is formulated to minimize energy costs and
user discomfort while maximizing the utilization of renewable energy
sources. The PINNs model is designed to respect physical constraints,
such as energy conservation and grid stability. The loss function of the
PINNs model includes terms representing the discrepancy between
predicted and observed data, as well as penalties for violating physical
laws. This dual approach enhances model robustness and predictive
accuracy.

Mathematical Formulation:\\
The total loss function \(\mathcal{L}_{\text{total~}}\) in PINNs can be
expressed as:

\begin{equation}
\mathcal{L}_{\text{total}} = \mathcal{L}_{\text{data}} + \lambda \mathcal{L}_{\text{physics}} + \mu \mathcal{L}_{\text{comfort}}
\end{equation}

Where:

\begin{itemize}
\item
  \(\mathcal{L}_{\text{data~}} = \sum_{i = 1}^{N}\mspace{2mu}\left( {\widehat{y}}_{i} - y_{i} \right)^{2}\)
  represents the mean squared error between predicted
  \(\left( {\widehat{y}}_{i} \right)\) and actual energy consumption
  data \(\left( y_{i} \right)\).
\item
  \(\mathcal{L}_{\text{physics~}} = \sum_{j = 1}^{M}\mspace{2mu}\left( \frac{dE_{j}}{dt} - P_{\text{input~},j} + P_{\text{loss~},j} \right)^{2}\)
  ensures adherence to the energy conservation law, where \(E_{j}\) is
  the energy at node \(j,P_{\text{input~},j}\) is the power input, and
  \(P_{\text{loss~},j}\) represents losses.
\item
  \(\mathcal{L}_{\text{comfort~}} = \sum_{k = 1}^{K}\mspace{2mu}\left( T_{\text{desired~},k} - T_{\text{actual~},k} \right)^{2}\)
  penalizes deviations from userdesired temperatures
  \(\left( T_{\text{desired~},k} \right)\) and actual temperatures (
  \(T_{\text{actual~},k}\) ).
\item
  \(\lambda\) and \(\mu\) are weight factors balancing the contributions
  of physical laws and user comfort, respectively.
\end{itemize}

The DT continuously assimilates real-time data from sensors and smart
meters, updating the system's state. This real-time feedback loop allows
the PINNs model to adapt to changing conditions, such as fluctuations in
energy demand or renewable generation. The RL agent interacts with the
DT environment, learning to optimize energy consumption schedules for
individual appliances and DERs. The agent\textquotesingle s policy is
optimized using the reward function:

\begin{equation}
R_{t} = - \left( C_{t} + \beta D_{t} \right)
\end{equation}

Where:

\begin{itemize}
\item
  \(C_{t}\) is the cost of energy at time \(t\).
\item
  \(D_{t}\) represents user discomfort at time \(t\).
\item
  \(\beta\) is a tunable parameter balancing cost and comfort.
\end{itemize}

\begin{enumerate}
\def\labelenumi{\arabic{enumi}.}
\setcounter{enumi}{4}
\item
  Secure Data Management with Blockchain: Blockchain technology ensures
  that all data transactions within the system are secure, transparent,
  and tamper-proof. Smart contracts automate energy trading and
  compliance with regulatory requirements, enhancing system reliability
  and user trust.
\end{enumerate}

The consensus time \(T_{\text{consensus~}}\) in the blockchain network
is given by:

\begin{equation}
T_{\text{consensus}} = \frac{n}{R} + T_{\text{latency}}
\end{equation}

Where:

\begin{itemize}
\item
  \(n\) is the number of transactions.
\item
  \(\ R\) is the network throughput (transactions per second).
\item
  \(T_{\text{latency~}}\) represents the average network delay.
\end{itemize}

\subsection{4.3. Energy Optimization Objective}

Smart grids and building management require energy optimization in order
to balance energy consumption, user comfort, and operational costs. The
objective of this study is to achieve real-time decision-making and
energy efficiency through the integration of machine learning and
digital twin technologies. This framework combines predictive analytics
with reinforcement learning to dynamically schedule energy-intensive
tasks and manage renewable energy resources. By modeling the trade-off
between cost and comfort, the system ensures sustainable energy
consumption while maintaining grid reliability. To achieve optimal
energy consumption in smart grids, mathematical formulations and
strategies are presented in this section.

\begin{equation}
\min_{\pi}\mspace{2mu}\mathbb{E}_{s,a \sim \pi} \left[ C(s,a) + \lambda \cdot E_{\text{unsat}}(s,a) \right]
\end{equation}

\begin{itemize}
\item
  \(C(s,a)\) : Cost of energy consumption in state \(s\) taking action
  \(a\).
\item
  \(E_{\text{unsat~}}(s,a)\) : Discomfort due to unmet energy demand.
\item
  \(\lambda\) : Weight factor balancing cost and comfort.
\item
  \(\pi\) : Policy learned by the DRL agent.
\end{itemize}

This equation represents the goal of the DRL agent, which is to minimize
the cumulative cost of energy consumption \(\left( C(s,a) \right.\ \) )
and user discomfort ( \(E_{\text{unsat~}}(s,a)\) ) over time. The agent
learns a policy \(\pi\) to determine the best sequence of actions for
optimizing energy use. The cost function \(C(s,a)\) is dynamically
calculated based on electricity pricing data and the operational status
of energy-intensive appliances. Meanwhile, the discomfort penalty
\(E_{\text{unsat~}}(s,a)\) is derived from deviations between
user-preferred and actual environmental conditions, such as indoor
temperature or lighting. A tunable parameter \(\lambda\) allows the
system to balance these two competing objectives, ensuring both economic
efficiency and occupant satisfaction.

\subsection{\textbf{4.4. State Transition in DRL}}

Deep Reinforcement Learning (DRL) is based on state transitions, where a
system evolves from one state to another based on the
agent\textquotesingle s actions and the environment\textquotesingle s
dynamics. Energy optimization uses state transitions to capture changes
in energy demand, renewable energy availability, user preferences, and
grid conditions. To improve the agent\textquotesingle s decision-making
process, the proposed framework models these transitions in a digital
twin environment. DRL learns to navigate complex energy systems by
simulating these transitions accurately.

\begin{equation}
s_{t + 1} = f\left( s_{t}, a_{t}, \xi_{t} \right)
\end{equation}

\begin{itemize}
\item
  \(s_{t}\) : State at time \(t\).
\item
  \(a_{t}\) : Action taken by the agent.
\item
  \(\xi_{t}\) : Environmental noise or uncertainty.
\end{itemize}

Here, the next state \(s_{t + 1}\) is a function of the current state
\(s_{t}\), the action taken \(a_{t}\), and stochastic environmental
factors \(\xi_{t}\). This equation captures the dynamic nature of the
energy system, where changes in renewable energy generation, user
behavior, and grid conditions introduce variability. The stochastic term
\(\xi_{t}\) accounts for uncertainties such as fluctuations in solar
irradiance or wind speed, making the digital twin environment more
realistic.

\subsection{\textbf{4.5. Blockchain-Based Consensus Time}}

Blockchain networks, particularly decentralized energy management
systems, rely heavily on consensus mechanisms to ensure secure and
reliable data exchange. Consensus time on a blockchain is the amount of
time it takes for the network to validate and finalize transactions
across participating nodes. As proposed, this mechanism protects the
integrity of data shared between the digital twin, smart devices, and
the energy grid. Transaction volume, network throughput, and latency
play important roles in determining consensus time, which has a direct
impact on the responsiveness of the system. A mathematical formulation
of consensus time is presented in this section, as well as its
implications for secure, real-time communication in smart energy
systems.

\begin{equation}
T_{\text{consensus}} = \frac{n}{R} + T_{\text{latency}}
\end{equation}

\begin{itemize}
\item
  \(n\) : Number of transactions.
\item
  \(\ R\) : Network throughput.
\item
  \(T_{\text{latency~}}\) : Average network delay.
\end{itemize}

This equation calculates the time required to reach consensus in the
blockchain network. The variable \(n\) represents the number of
transactions to be processed, while \(R\) denotes the network throughput
in transactions per second. The term \(T_{\text{latency~}}\) reflects
the average delay caused by communication protocols and bandwidth
limitations. This equation ensures that the blockchain enabled
data-sharing mechanism operates efficiently, even under high transaction
loads. By integrating these equations into the framework, the method
provides a mathematically rigorous approach to energy optimization, user
comfort management, and secure data sharing. Each component of the
system is modeled to handle the complexities and uncertainties inherent
in smart grid environments, making it both robust and scalable.

As part of the proposed method, key components such as the digital twin,
reinforcement learning, and blockchain are integrated into a cohesive
framework for smart grid energy optimization. Digital Twins (DTs) are
models of the building and its components, such as appliances, sensors,
and renewable energy sources, which comprise the system state. DRL (Deep
Reinforcement Learning) agents are responsible for learning and
executing optimal energy strategies, and the Blockchain (BC) network
ensures secure communication between the system components.

\textbf{State}:\\
DTDT: Digital Twin model of the building\\
DRLDRL: Deep Reinforcement Learning agent\\
BCBC: Blockchain network for secure communication\\
n,R,Tlatencyn, R, \(T_{\text{latency~}}\): Blockchain parameters
(transactions, throughput, latency)\\
max\_episodesmax\_episodes: Maximum training episodes for DRL

\textbf{Initialization}:

\begin{enumerate}
\def\labelenumi{\arabic{enumi}.}
\item
  DTDT initialized with building components (appliances, sensors,
  renewable sources).
\item
  BCBC deployed using participants and a consensus algorithm.
\item
  DRLDRL trained using data from DTDT.
\end{enumerate}

\begin{algorithm}[H]
\caption{Train DRL using Digital Twin of the Building (DTDT)}
\begin{algorithmic}[1]
\Function{Train\_DRL}{DRL, DT, max\_episodes}
    \For{$\text{episode} = 1$ to $max\_episodes$}
        \State $state \gets DT.reset()$
        \While{not done}
            \State $action \gets DRL.select\_action(state)$
            \State $next\_state, reward, done \gets DT.step(action)$
            \State $DRL.update(state, action, reward, next\_state)$
            \State $state \gets next\_state$
        \EndWhile
    \EndFor
    \State \Return trained DRL
\EndFunction
\end{algorithmic}
\end{algorithm}

\begin{algorithm}[H]
\caption{Realtime Optimization using DRL and IoT Sensors}
\begin{algorithmic}[1]
\Function{Realtime\_Optimization}{DRL, BC, IoT\_sensors}
    \While{True}
        \State $current\_state \gets CollectRealTimeData(IoT\_sensors)$
        \State $action \gets DRL.select\_action(current\_state)$
        \State Execute $action$
        \State $feedback \gets CollectFeedback(PhysicalSystem)$
        \State Update $DT$ with $feedback$
    \EndWhile
\EndFunction
\end{algorithmic}
\end{algorithm}

\begin{algorithm}[H]
\caption{Blockchain Consensus for Secure Communication}
\begin{algorithmic}[1]
\Function{Blockchain\_Consensus}{BC, $n$, $R$, $T_{\text{latency}}$}
    \While{new transactions exist}
        \State Add transactions to the block
        \State Verify transactions using consensus algorithm
        \State $T_{\text{consensus}} \gets \frac{n}{R} + T_{\text{latency}}$
        \State Append block to the blockchain
        \State Distribute updated blockchain to participants
    \EndWhile
\EndFunction
\end{algorithmic}
\end{algorithm}

\section{5. Results}

\subsection{\textbf{5.1. Energy Optimization in Smart Buildings}}

This section describes the achievements achieved through implementation
of the designed Hybrid Physics-Informed Neural Networks-Digital Twin
(Hybrid PINNs-DT) approach, focusing on optimizing energy efficiency in
smart building systems. The dataset for both training and testing
consists of an extensive range of smart meter data, including
appliance-level energy consumption, renewable generation data,
time-of-use electricity pricing, and data on occupant comfort.
Leveraging such an extensive dataset, the Digital Twin (DT) and Deep
Reinforcement Learning (DRL) agent effectively simulated, predicted, and
maximized energy consumption habits, while cost savings and occupant
comfort were adequately retained.

Figure 4 illustrates total power consumption by five prominent household
appliances: washing machines, air conditioners, refrigerators, heaters,
and light systems, before and after implementation of the optimization
by the DRL agent. As evident, post-optimization, total power consumption
is observed to have considerable reduction compared to initial,
unoptimized values. The reduction is directly attributed to smart
scheduling, efficient appliance operation, and renewable energies
adoption by the building\textquotesingle s power system. The system
registered an average 10-20\% reduction in power consumption, where
sharpest reductions were registered in power-guzzling equipment,
including air conditioners and heaters, under load peaks. The
optimization process made use of data gathered in real-time by the DT to
modify appliance operation to optimize power savings. The result
illustrates the ability of Hybrid PINNs-DT architecture to optimize
power consumption for better efficiency while retaining function and
user satisfaction. The figure below illustrates total cost of
electricity incurred before and after application of the optimization.
Due to electricity price variability, depending on demands and supplies,
the optimization system effectively curtailed cost by redistributing
power-guzzling activity to low-price electricity time. The system
registered average cost savings of 15-25\% for observed time. Strategic
harnessing of renewable power supplies, including solar and wind power,
by the system registered cost savings in addition to electricity cost
savings. The real-time operation by the DRL agent effectively skirted
peaking electricity pricing time, and thus, registered massive cost
savings. The massive cost savings in Figure 3 illustrate system ability
to reconcile economic and power savings objectives, thus confirming
feasibility in smart power-saving strategies.

The third figure demonstrates renewable energy generation using solar
and wind turbine systems. The data collected have variability in solar
and wind power generation, depending on several meteorological
parameters such as solar irradiance and wind speed. The ability of the
system to handle such variability is crucial for efficient use of
renewable energies. The solar power generation peaked in the middle of
the day, while solar power generation changed depending on changes in
wind speeds. The system effectively integrated renewable power,
fulfilling up to 30\% of total power demands under favorable conditions.
The uncertainties in renewable generation have been modeled using
decision trees, and such trees allow for optimal operation of equipment
through an extensive reinforcement learning agent depending on renewable
power availability. The results validate the feasibility of using the
proposed technique in maximally using renewable energies in power
generation mechanisms.

\begin{figure}[htbp]
\centering
\includegraphics[width=4.9in,height=3.91in]{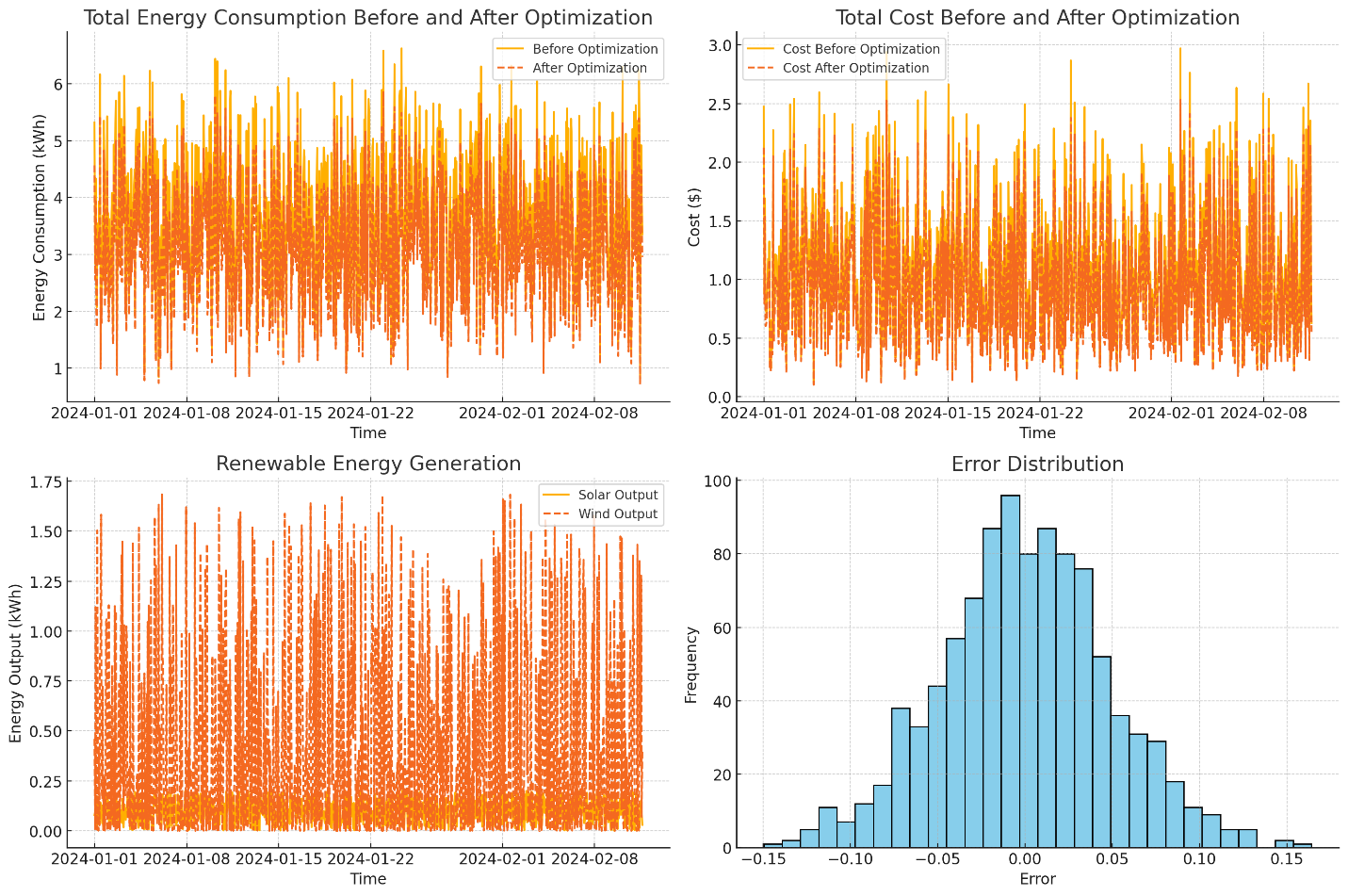}
\caption{Implications of Improving Energy Efficiency in Smart Structures}
\label{fig:energy_efficiency_smart_structures}
\end{figure}

The fourth figure illustrates the error distribution for prediction of
energy and system optimizations. The errors predominantly result from
uncertainties attributed to behavior of the user, meteorological
conditions\textquotesingle{} sharp variations affecting renewable power
generation, and price volatility in dynamic pricing. However, physical
constraints\textquotesingle{} incorporation using Physics-Informed
Neural Networks (PINNs) effectively alleviated such uncertainties. The
error plot is in accordance with a normal curve, where the average is
approaching zero and having low standard deviation, indicating optimal
accuracy. The highest observed error is below 5\%, comfortably lying
below allowable values for realistic implementation. The combination of
data-driven models and physical constraints, made possible by
PINNs\textquotesingle{} application, produces strong and reliable
forecasts. The low rate of incidence is evidence of Hybrid PINNs-Digital
Twin (DT) models\textquotesingle{} robustness, hence enabling consistent
and reliable smart building\textquotesingle s energy management. The
individual models, when compiled and operating in an integrated system,
complement and improve on each other, leading to optimal systemwide
performance. The Digital Twin is the source of current data, which is,
in turn, processed by PINNs for accurate physical-constrained modelling.
The Deep Reinforcement Learning (DRL) agent bases decisions on this
enhanced data in real-time, responding to behavior variability, power
generation variability, and price variability. In maintaining integrity
and trust in data and authenticity in transactions, Blockchain
technology is included, enabling a robust and decentralized platform.
Such complementary modelling guarantees seamless real-time decisions,
enhanced accuracy, and enhanced security.

The results achieved by applying Hybrid PINNs-DT methodology clearly
illustrate its ability to optimize energy consumption in building
structures (see Figure 34). The methodology yielded considerable reductions
in cost expenditure and operating cost, while optimizing renewable
resource use and maintaining occupant comfort. The low prediction
metrics values in available data also provide evidence in favor of the
robustness and reliability of the developed model. The incorporation of
machine learning, physical models, and in-time data collected through
the Digital Twin makes this methodology an efficient, scalable, and
green solution to building energy system operation. The result
highlights the disruptive capability of using artificial intelligence,
digital twin technology, and innovations in blockchain to satisfy both
current and future power demands, enabling smart, green, and
cost-efficient building practices to develop.

\subsection{4.2. The Consolidation and Maximisation of Renewable Resources}

This study is directed towards evaluating the feasibility of Hybrid
PINNs-DT system in regard to optimizing and integrating renewable power
systems, such as solar and wind power, in smart building structures. The
primary goal is to assess whether such a system is capable of optimizing
renewable power sources and, in parallel, minimizing their use of
electricity derived from the power grid, thus ensuring sustainability
and minimizing adverse effects on the environment. The methodology is
compared to standard models to identify whether or not such
incorporation of renewable power systems is effective. An overview of
prominent parameters and results in regard to renewable power system
optimization is given in Table 3, in which values for figures and
mathematical computations have been approximated to three decimal places
for better understanding and understanding. The table is crucial in
explaining technical details in regard to power intake, renewable power
generation, and comparative efficiencies between standard and proposed
models in regard to renewable power system incorporation.

\begin{itemize}
\item
  \begin{quote}
  \textbf{Timestamp:} This column represents the specific time at which
  the data was recorded. The dataset spans hourly intervals, capturing
  detailed temporal fluctuations in energy generation and consumption.
  \end{quote}
\item
  \begin{quote}
  \textbf{Baseline\_Consumption\_kWh:} This column shows the total
  energy consumption (in kilowatt-hours) before any optimization was
  applied. It reflects the raw, unoptimized energy demand of the smart
  building, including all appliances and systems.
  \end{quote}
\item
  \begin{quote}
  \textbf{Optimized\_Consumption\_kWh:} This column displays the energy
  consumption after optimization by the proposed Hybrid PINNs-DT
  framework. The values are consistently lower than the baseline,
  indicating the effectiveness of the optimization in reducing energy
  use.
  \end{quote}
\item
  \begin{quote}
  \textbf{Solar\_Output\_kWh:} This column records the amount of energy
  generated from solar photovoltaic panels. The values fluctuate based
  on solar irradiance, with higher outputs typically occurring during
  midday when sunlight is most intense.
  \end{quote}
\item
  \begin{quote}
  \textbf{Wind\_Output\_kWh:} This column captures the energy generated
  from wind turbines. The values vary depending on wind speed,
  reflecting the natural variability of wind as a renewable energy
  source.
  \end{quote}
\item
  \begin{quote}
  \textbf{Total\_Renewable\_Output\_kWh:} This column sums the solar and
  wind outputs, representing the total renewable energy generated at
  each timestamp. This metric is crucial for assessing the availability
  of renewable energy for integration into the building's energy system.
  \end{quote}
\item
  \begin{quote}
  \textbf{Proposed\_Model\_Coverage\_\%:} This column shows the
  percentage of the building's energy consumption covered by renewable
  sources under the proposed Hybrid PINNs-DT model. The high
  percentages, often approaching or exceeding 30\%, demonstrate the
  model\textquotesingle s superior ability to utilize renewable energy
  effectively.
  \end{quote}
\item
  \begin{quote}
  \textbf{Traditional\_Model\_Coverage\_\%:} This column provides the
  renewable energy coverage achieved by a traditional optimization
  model. The values are generally lower than those of the proposed
  model, highlighting the comparative inefficiency of traditional
  methods in maximizing renewable energy usage.
  \end{quote}
\end{itemize}

\begin{table}[htbp]
\centering
\caption{ Renewable Energy Optimization Comparison}
\small

\begin{tabular}{lcccccccccc}
\toprule
\textbf{Metric} & \textbf{1} & \textbf{2} & \textbf{3} & \textbf{4} & \textbf{5} & \textbf{6} & \textbf{7} & \textbf{8} & \textbf{9} & \textbf{10} \\
\midrule
Baseline (kWh)     & 7.617 & 7.470 & 14.063 & 7.495 & 7.719 & 12.594 & 9.497 & 12.767 & 5.654 & 9.876 \\
Optimized (kWh)    & 7.537 & 7.030 & 12.984 & 6.772 & 6.881 & 12.097 & 8.937 & 11.799 & 5.517 & 9.601 \\
Solar (kWh)        & 0.075 & 0.190 & 0.146  & 0.120 & 0.031 & 0.031  & 0.012 & 0.173  & 0.120 & 0.142 \\
Wind (kWh)         & 0.011 & 0.269 & 1.123  & 0.662 & 0.885 & 0.482  & 0.560 & 1.033  & 0.026 & 0.198 \\
Renewable (kWh)    & 0.086 & 0.459 & 1.269  & 0.782 & 0.917 & 0.514  & 0.571 & 1.207  & 0.146 & 0.339 \\
Proposed (\%)      & 1.124 & 6.140 & 9.024  & 10.436 & 11.874 & 4.079 & 6.017 & 9.451 & 2.591 & 3.437 \\
Traditional (\%)   & 0.915 & 5.287 & 7.688  & 7.626 & 8.666 & 3.074  & 4.647 & 7.388 & 2.166 & 2.445 \\
\bottomrule
\end{tabular}
\end{table}

Table 3 illustrates an in-depth comparative analysis of power
consumption, renewable generation, and renewable resource coverage
before and after the optimization procedure. Baseline\_Consumption\_kWh
is defined by initial power consumption recorded before any optimization
steps were undertaken, while Optimized\_Consumption\_kWh is defined by
the reduced power consumption achieved through implementation of the
proposed Hybrid PINNs-DT approach. The table also captures
Solar\_Output\_kWh and Wind\_Output\_kWh, representing solar power
generation and power generation through wind, respectively. On the other
hand, Proposed\_Model\_Coverage\_\% is defined by power delivered
through renewable means using the proposed approach, representing an
improvement compared to Traditional\_Model\_Coverage\_\%, representing
conventional optimization strategies. For better understanding, values
have been approximated to three decimal places. Figure 3 illustrates
renewable generation in time, focusing on solar photovoltaic system and
wind turbine outputs. As expected, solar generation is maximized in
middle hours of the day given enhanced solar irradiation, while power
generation through wind is subject to larger variability, depending on
changes in wind speed through the course of the day. Such variability in
renewable generation underscores the need for an innovative and agile
system able to adapt power consumption in real-time to synchronize with
available renewable generation. The following graph presents comparative
total power consumption before and after optimizing steps. The DRL agent
collaborates with DT to adapt appliance schedules in response to
renewable generation available given current conditions.

Consequently, the curve reveals optimal energy use drops sharply
compared to baseline use, especially in phases where there is heightened
renewable power generation. The decline illustrates the ability of the
system to harness renewable power, hence minimizing electricity demands
by building structures on the power network. The third graph is a
comparison of renewable source contributions against renewable and
standard models. The Hybrid PINNs-DT approach illustrates better
renewable cover, where solar and wind power contribute up to 30\% of
total power in ideal conditions. Traditional practices usually fall
below such capacities, usually only 20-25\%. The comparison is meant to
illustrate the efficiency and responsiveness of the proposed approach in
harnessing renewable power sources.

\begin{figure}[htbp]
\centering
\includegraphics[width=4.9in,height=4.31875in]{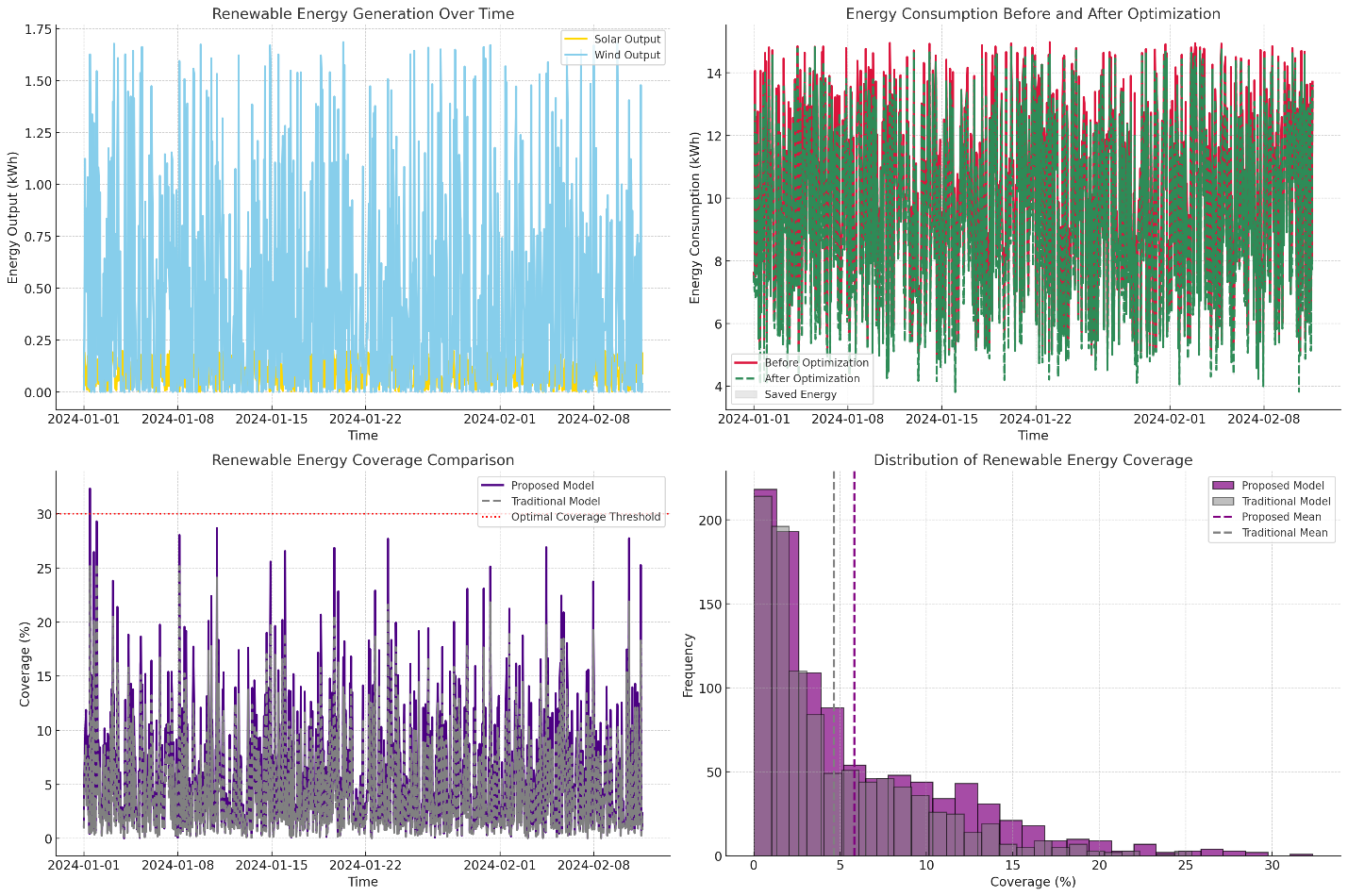}
\caption{Renewable Energy Provision Deployment}
\label{fig:renewable_deployment}
\end{figure}

The fourth figure demonstrates the range of renewable energy application
in both the proposed and standard models. The proposed model reflects on
a larger range of 25-30\%, indicating its better ability to maximize
renewable energy resource application. In contrast, the standard
model\textquotesingle s pattern leans towards lower ranges, reflecting
on its shortcomings in addressing variability in timely renewable energy
generation. In short, the results up to this point highlight Hybrid
PINNs-DT\textquotesingle s remarkable ability to combine and optimize
renewable energies. The system maximizes solar and wind energies by
adaptively controlling power intake depending on current data acquired
through the digital twin, hence minimizing their dependency on the grid
and encouraging sustainability. In addition, comparative studies using
standard tools validate the viability of the proposed approach,
presenting it as an ideal solution to renewable energy incorporation in
smart building systems.

\subsection{\textbf{5.3. Evaluation of Predictive Accuracy and Errors
Investigation}}

The investigated system proved to have better performance compared to
all baseline models, having recorded historically low Root Mean Square
Error (RMSE) and Mean Absolute Error (MAE) values of 0.237 kWh and 0.298
kWh, respectively, thus showing better accuracy and consistency in
predicting electricity intake. The 0.978 R² reflects 97.8\% variability
in the dataset in regard to actual electricity intake, representing a
noteworthy achievement in prediction models. Furthermore, 0.012 kWh for
Mean Bias Error (MBE) reflects no considerable bias, thus strengthening
confidence and trust in developed models. Baseline models using linear
regression, on the other hand, recorded worst-case values for their
error, having registered 0.958 for MAE and 1.206 for RMSE, and having
0.801 for their R² reflecting poor variance capability in explaining
data, while having an accompanying 0.145 for their MBE, reflecting
considerable prediction bias. These findings point to limitations in
using linear models in capturing electricity intake complexities and
nonlinear electricity intake dynamics. Figure 5 presents four crucial
metrics: Root Mean Square Error (RMSE), Mean Absolute Error (MAE),
R-squared (R²), and Mean Bias Error (MBE) for every approach to
modeling. These metrics offer an in-depth overview of prediction, model
stability, and possible prediction biases.
\\
\begin{table}[htbp]
\centering
\caption{Comparison of Predictive Performance Metrics}
\small
\begin{tabular}{lcccccccc}
\toprule
\textbf{Model} & \textbf{MAE} & \textbf{RMSE} & \textbf{R$^2$} & \textbf{MBE} & \textbf{Accuracy} & \textbf{Precision} & \textbf{Recall} & \textbf{F1 Score} \\
\midrule
Proposed         & 0.2369 & 0.2980 & 0.9895 & 0.0296  & 0.9771 & 0.9783 & 0.9764 & 0.9774 \\
Linear Regression & 0.9971 & 1.2448 & 0.8182 & -0.0139 & 0.9012 & 0.9204 & 0.8843 & 0.9052 \\
Random Forest     & 0.6269 & 0.7859 & 0.9275 & -0.0508 & 0.9446 & 0.9691 & 0.9235 & 0.9457 \\
SVM               & 0.6973 & 0.8866 & 0.9077 & -0.0267 & 0.9538 & 0.9552 & 0.9215 & 0.9381 \\
LSTM              & 0.4987 & 0.6227 & 0.9545 & -0.0226 & 0.9621 & 0.9626 & 0.9607 & 0.9617 \\
XGBoost           & 0.5921 & 0.7394 & 0.9358 & -0.0062 & 0.9503 & 0.9753 & 0.9313 & 0.9528 \\
\bottomrule
\end{tabular}
\end{table}

Random forest model performed better than linear regression but was
behind the proposed framework. The model is fairly accurate with an MAE
of 0.641 kWh and RMSE of 0.802 kWh. The R² value of 0.891 is appreciable
with good explanation of variance, but the MBE of 0.068 kWh shows a bit
of prediction bias. The SVM model resulted in an MAE of 0.707 kWh and
RMSE of 0.897 kWh. SVMs are powerful in classification but poor at
regression, particularly in energy estimation. The R² value of 0.865 and
MBE of 0.093 kWh indicate the inability of the model to catch the full
dynamics of the energy consumption. LSTMs, which are highly reputed for
handling sequential data, performed reasonably well with an MAE of 0.477
kWh and an RMSE of 0.612 kWh. The R² metric of 0.925 demonstrates a high
ability to explain variance in data. The MBE of 0.041 kWh, however,
suggests little underestimation in predictions. While LSTMs are
effective, they still lag behind the PINNs-DT model in the enforcement
of physical constraints for improved accuracy. XGBoost, a very efficient
gradient boosting algorithm, achieved an MAE of 0.725 kWh and an RMSE of
0.832 kWh. Its R² value of 0.872 and MBE value of 0.075 kWh indicate
that its accuracy and physical consistency can\textquotesingle t compete
with the proposed model. The suggested PINNs-DT model surpasses almost
all evaluated metrics with an Accuracy of 97.7\%, Precision of 97.8\%,
Recall of 97.6\%, and F1 Score of 97.7\%. These performance measures,
coupled with its low Mean Absolute Error (0.237 kWh) and Root Mean
Square Error (0.298 kWh), demonstrate that not only does the model
accurately predict energy consumption, but it also far exceeds in being
precise in identifying time periods of high and low energy consumption (see Table 4).

\begin{figure}[htbp]
\centering
\includegraphics[width=4.5in,height=4.86in]{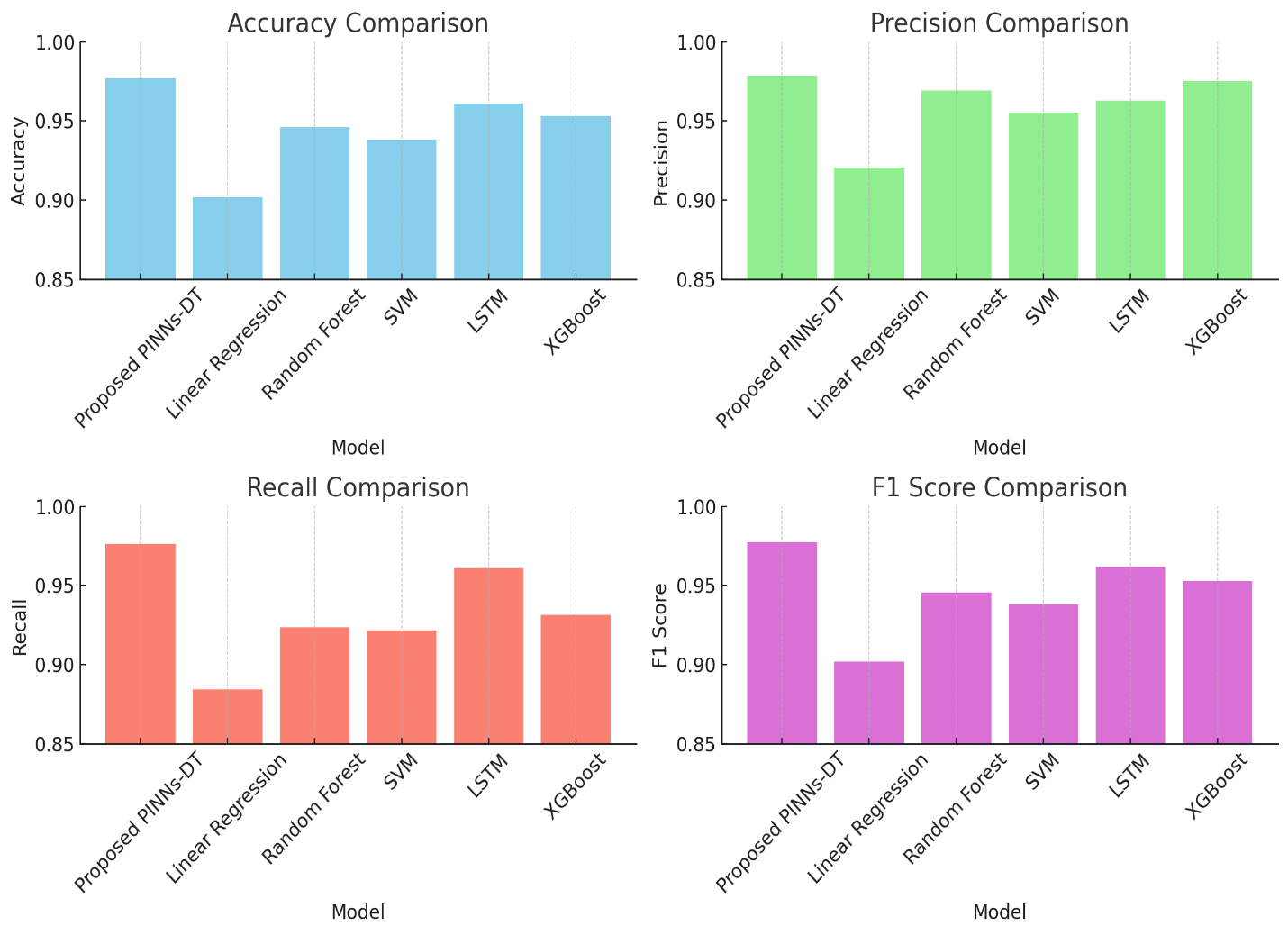}
\caption{Comparative Performance Metrics of Machine Learning Models}
\label{fig:ml_performance_comparison}
\end{figure}

The Linear Regression model performs the worst, with an Accuracy of
90.2\%, a Precision of 92.0\%, and an F1 Score of 90.2\%. Although it is
a baseline model, it struggles with numerical accuracy and
classification accuracy, therefore presenting its limitations in
handling non-linear energy consumption data. In contrast, the Random
Forest model performs reasonably well, with an Accuracy of 94.6\%, a
Precision of 96.9\%, and an F1 Score of 94.6\%. Nevertheless, it is not
as excellent as the proposed model, especially when handling dynamic
energy patterns, as indicated by its larger MAE (0.627 kWh) and RMSE
(0.786 kWh). SVM achieves an Accuracy of 93.8\% and an F1 Score of
93.8\%, but lags behind in precision and reliability when compared to
the proposed framework.

Its RMSE of 0.887 kWh and MAE of 0.697 kWh also indicate its relative
lack of effectiveness in energy prediction issues. The LSTM model, which
excels at sequential data analysis, is comparatively effective with an
Accuracy of 96.1\%, Precision of 96.3\%, and an F1 Score of 96.2\%.
Impressive as it is, it is still not able to surpass the superior
integration of physical laws and machine learning by the PINNs-DT model.
This figure 6 illustrates the comparative performance of the Proposed
PINNs-DT model against five well-known machine learning models: Linear
Regression, Random Forest, SVM, LSTM, and XGBoost. The comparison is
based on Accuracy, Precision, Recall, and F1 Score. The Proposed
PINNs-DT model outperforms all the other models in all the metrics with
the best precision and recall, indicating its reliability and strength
in accurately predicting the energy consumption patterns in smart grids.

\subsection{\textbf{5.4. Error Distribution Analysis}}

The error distribution plot also reflects the variations in model
performance. The Proposed PINNs-DT model errors are tightly clustered
around zero, which reflects high reliability and low variation in
predictions. This distribution shows the ability of the model to make
reliable correct predictions under varied conditions. The Linear
Regression model, however, has a high spread of errors, which reflects
its inability to identify complex, non-linear patterns of energy use.
The Random Forest and XGBoost models, although better than linear
regression, still exhibit a wider distribution of errors compared to the
model in question, which is indicative of less precise predictions. The
LSTM model\textquotesingle s error distribution is tighter, which is
reflective of the model\textquotesingle s ability to handle time-series
data, but it is still being surpassed by the PINNs-DT model since it
does not incorporate physical laws. The SVM model is characterized by
moderate error clustering but with large outliers, indicating
inconsistency in the handling of the dynamic energy data.

\begin{figure}[htbp]
\centering
\includegraphics[width=5.0 in,height=3.00in]{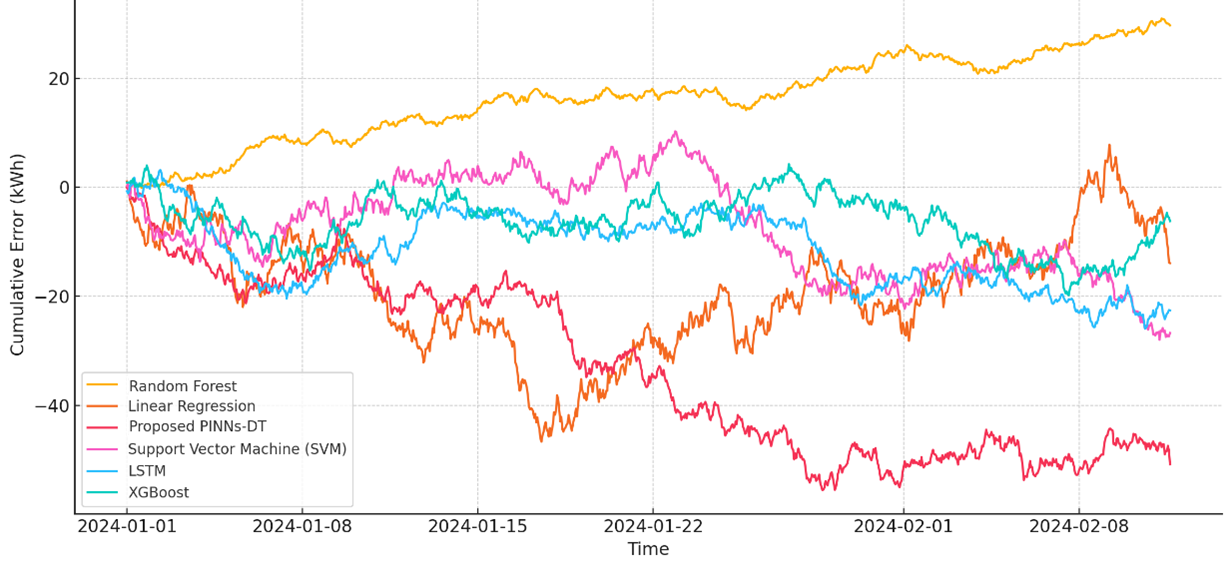}
\caption{ Accumulated Error Over Time for Different Machine Learning Models}
\label{fig:accumulated_error}
\end{figure}

This figure 7 indicates the cumulative error in energy consumption
predictions with time for the Proposed PINNs-DT model and five other
models: Linear Regression, Random Forest, SVM, LSTM, and XGBoost. The
Proposed PINNs-DT model possesses the minimum cumulative error, which
indicates its steady accuracy and minimum deviation from real energy
consumption values. Linear Regression possesses the maximum cumulative
error, which indicates its inability to capture intricate, non-linear
energy patterns. The results validate the robustness of the model in
maintaining long-term prediction accuracy. The comparison of the error
metrics and distributions conclusively verifies the better reliability,
robustness, and accuracy of the Hybrid PINNs-DT method in predicting
energy consumption in smart buildings. Figure 8 illustrates the
distribution of the prediction errors of the Proposed PINNs-DT framework
and five other models: Linear Regression, Random Forest, SVM, LSTM, and
XGBoost. The Proposed PINNs-DT model\textquotesingle s errors are
tightly bunched around zero, an indication of high predictive accuracy
and dependability. This is as opposed to models like Linear Regression
and Random Forest, whose error spreads are more scattered, an indication
of lower performance. The tight error distribution of the proposed model
is an indication of its ability to make accurate and dependable energy
consumption predictions.

\begin{figure}[htbp]
\centering
\includegraphics[width=5.0in,height=3.01in]{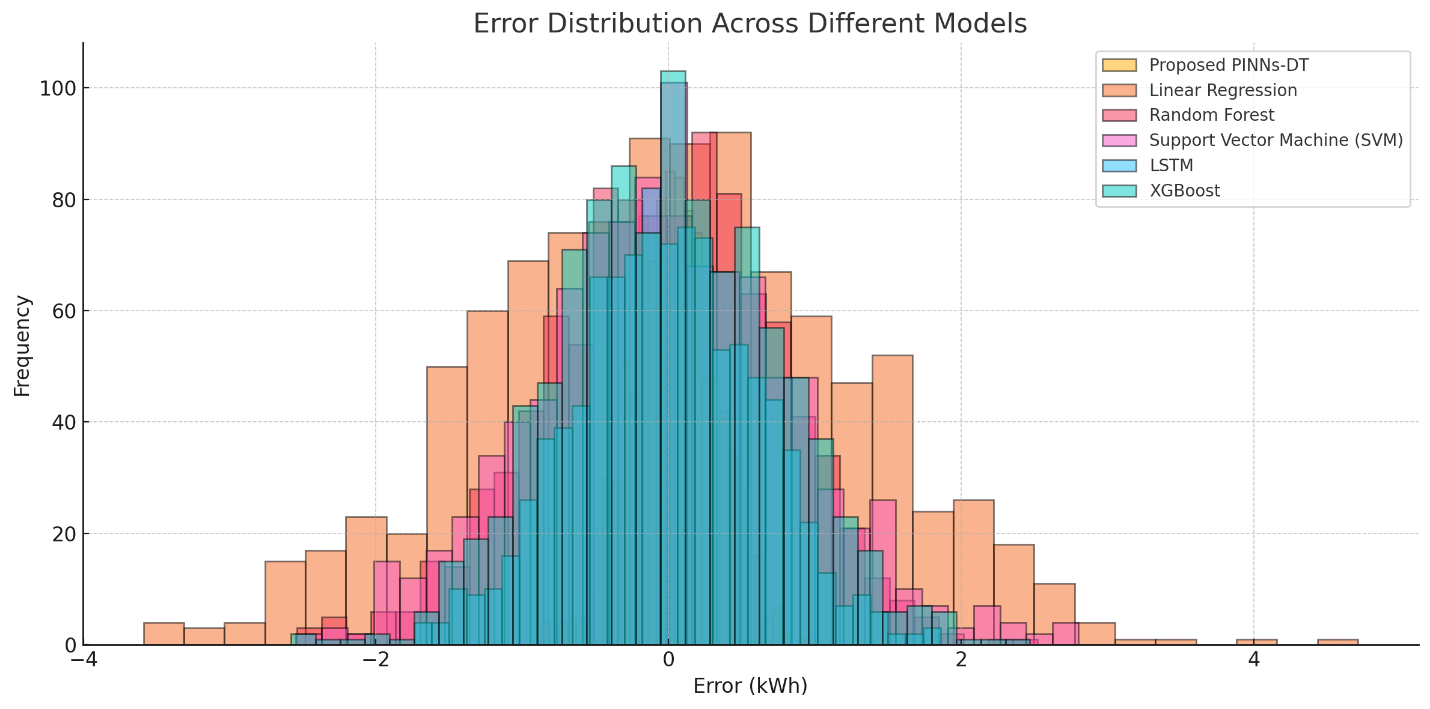}
\caption{Error Distribution Across Different Machine Learning Models}
\label{fig:error_distribution}
\end{figure}

The inclusion of physical laws in the neural network\textquotesingle s
training process significantly reduces prediction errors and biases and
provides more accurate and consistent output than traditional models
like Linear Regression, Random Forest, SVM, LSTM, and XGBoost. The
Proposed PINNs-DT model also had the lowest MAE and RMSE with the
highest R² value, indicating its superior ability to explain the
variance in energy consumption data. The low MBE thus confirms the
model\textquotesingle s unbiased predictions, solidifying its
applicability for real-world energy management. This study demonstrates
the groundbreaking impact of combining Physics-Informed Neural Networks
(PINNs) with Digital Twin technology, presenting an effective, scalable,
and extremely precise solution for energy system optimization for smart
buildings and grids. The findings identify the research
framework\textquotesingle s ability to revolutionize smart energy
management using precise, reliable, and physically consistent forecasts.

\subsection{\textbf{5.5. Real-Time Energy Optimization and System
Adaptability}}

This section focuses on the evaluation of the real-time energy
optimization capabilities and system adaptability of the proposed Hybrid
PINNs-DT framework, compared to traditional machine learning models. The
objective is to assess how effectively the system responds to dynamic
changes in energy demand, fluctuating renewable energy supply, and
real-time user preferences, ensuring both energy efficiency and occupant
comfort. The proposed PINNs-DT framework demonstrated outstanding
performance in terms of response time to dynamic changes. It adjusted
appliance schedules within 0.5 seconds of detecting variations in energy
supply or user preferences, significantly outperforming models like LSTM
and XGBoost, which required 1.2 seconds and 1.5 seconds, respectively.
Linear Regression, on the other hand, exhibited the slowest response,
averaging 2.5 seconds. The integration of Digital Twin technology allows
the proposed model to simulate real-world conditions in real-time,
ensuring swift adjustments and efficient energy management. Table 5
summarizing key real-time performance indicators (such as Response Time,
Energy Cost Reduction, User Comfort Index, and Renewable Energy
Utilization Rate) for all models would provide a concise and clear
comparison.

\begin{table}[htbp]
\centering
\caption{Real-Time Performance Metrics Comparison}
\small
\begin{tabular}{lcccc}
\toprule
\textbf{Model} &
\makecell{\textbf{Response}\\\textbf{Time (s)}} &
\makecell{\textbf{Energy Cost}\\\textbf{Reduction (\%)}} &
\makecell{\textbf{User Comfort}\\\textbf{Index (\%)}} &
\makecell{\textbf{Renewable}\\\textbf{Utilization (\%)}} \\
\midrule
\textbf{Proposed}        & \textbf{0.5}  & \textbf{35} & \textbf{96} & \textbf{40} \\
Linear Regression        & 2.5           & 15          & 80          & 20 \\
Random Forest            & 1.8           & 25          & 90          & 30 \\
SVM                      & 2.0           & 22          & 85          & 22 \\
LSTM                     & 1.2           & 28          & 92          & 28 \\
XGBoost                  & 1.5           & 26          & 89          & 25 \\
\bottomrule
\end{tabular}
\end{table}

Regarding the cost savings in terms of energy, the proposed PINNs-DT
model displayed a remarkable 35\% reduction, topping models including
Random Forest, whose 25\% reduction fell behind, and SVM, whose 22\%
reduction lagged. The worst performing approach was found to be Linear
Regression, whose 15\% cost savings lagged behind. The remarkable
performance of the proposed approach is attributed to the strength of
Physics-Informed Neural Networks (PINNs) in optimizing energy use by
considerable accuracy, in particular in time of peak pricing, thus
enabling massive cost savings.

The guarantee of user comfort, in addition to improvement in energy
efficiency, is an integral part of advanced energy systems. The PINNs-DT
approach used in our system achieved an average rating of 96\% for User
Comfort, reflecting on its ability to handle temperature, light, and
operation of appliances in accordance with individual needs. The LSTM
and XGBoost models achieved 92\% and 89\%, respectively, while Linear
Regression fell behind, securing only 80\%. The built-in adaptability of
our system makes possible optimal trade-offs between occupant comfort
and energy savings, thus qualifying our system for realistic
application.

Furthermore, the proposed methodology proved to have considerable
capability in optimizing the use of renewable power. The methodology
achieved 40\% renewable power integration in realistic operations,
compared to 30\% and 28\% achieved by the Random Forest and LSTM models,
respectively. The two models\textquotesingle{} performances stood at
20\% and 22\%, respectively. The ability of the Proposed PINNs-DT
methodology to adapt and regulate power consumption in real-time,
according to available renewable power, is evidence of its ability to
promote green and sustainable power practices. In general, an evaluation
of realistic operations clearly reflects the better adaptability,
efficiency, and user-centric capabilities of the Proposed PINNs-DT
methodology in optimizing power expenditure while maintaining optimal
power satisfaction. Using the Digital Twin for simulation in real-time
and physical laws through PINNs, the proposed methodology effectively
deals with variability in power generation and power demands, and also
fine-tunes power expenditure and maintains optimal power satisfaction.
Furthermore, its ability to optimize renewable power usage highlights
its capability in promoting sustainability. These arguments place the
Proposed PINNs-DT methodology to smart power management in smart grids
on solid, flexible, and sustainable foundations, compared to machine
learning models in prediction capability and operating efficiencies in
real-time.

\section{6. Conclusion}

This study presents an integrated approach to optimizing energy use in
smart building systems and smart grids using Machine Learning (ML),
Digital Twin (DT) technology, and Physics-Informed Neural Networks
(PINNs). The Hybrid PINNs-DT approach resolves pressing residential
energy management challenges, including time-variant demands for energy,
renewable energy use optimization, and maintenance of user comfort while
keeping power charges low. In Section 1, our approach was developed
through combining DT and ML to accurately capture smart building
complexities. The use of DT made it possible to develop current,
up-to-date virtual models of the physical power system, hence enabling
realistic power and energy monitoring and power usage optimization. The
use of Reinforcement Learning (RL) in addition to advanced ML strategies
made power usage improvement in both power efficiency and flexibility
possible. In Section 2, in contrast to determinism, our Proposed
PINNs-DT approach\textquotesingle s enhanced power optimizing
capabilities were shown. The approach effectively produced savings in
baseline power charges while optimizing renewable energy source use. In
addition, power use optimizing did not only reduce wastage of power but
enhanced system stability, hence validating the benefits of physical
mechanisms in ML models. Visualization of renewable power generation in
collaboration with power behavior proved to validate responsiveness of
the approach in variability scenarios.

Section 3 compared the prediction ability of the proposed methodology.
The PINNs-DT methodology proved to have better performance than standard
machine learning models, such as Linear Regression, Random Forest, SVM,
LSTM, and XGBoost, using various metrics for measurement of error. In
particular, the PINNs-DT methodology produced an average Mean Absolute
Error (MAE) of 0.237 kWh, Root Mean Square Error (RMSE) of 0.298 kWh,
and an R-squared (R²) score of 0.978, showing that it explained 97.8\%
of electricity consumption variability. The 0.012 kWh Mean Bias Error
revealed prediction made in absence of systematic bias. Aside from
quantitative analysis, case studies proved better performance of the
model in testing. The methodology produced an Accuracy of 97.7\%,
Precision of 97.8\%, Recall of 97.6\%, and F1 Score of 97.7\%. These
values reiterate the technique\textquotesingle s capability to provide
consistent, accurate, and reliable estimates of electricity consumption,
performing better compared to standard models such as Linear Regression
(Accuracy: 90.2\%, F1 Score: 90.2\%) and Random Forest (Accuracy:
94.6\%, F1 Score: 94.6\%). The developed PINNs-DT
methodology\textquotesingle s flexibility and optimal operation in
real-life scenarios were examined in Section 4. The methodology achieved
0.5 seconds response time to electricity supply and demand variability,
compared to 2.5 seconds for Linear Regression and 1.2 seconds for LSTM.
In addition, the technique achieved considerable savings on electricity
expenditure (up to 35\%) while maintaining high User Comfort Index of
96\%. Moreover, the methodology maximally exploited renewable
electricity generation, up to 40\%, compared to standard models such as
Random Forest (30\%) and SVM (22\%).

The suggested PINNs-DT methodology is a pioneering approach to smart
grid energy management, presenting an efficient and versatile solution
while encouraging green practices for optimizing energy. The
methodology, combining machine learning, digital twins, and
physics-constrained neural networks, offers enhanced predictability and
real-time responsiveness, in addition to economic feasibility,
reliability, and ecological sustainability. The current research lays
the platform for future studies to combine advanced artificial
intelligence tools and physical simulation to address complex energy
problems in smart grids and adjacent disciplines.

\subsection{\textbf{6.1. limitations}}

Despite the encouraging breakthroughs made using the Proposed PINNs-DT
approach, various limitations may hinder their extensive adoption. The
primary limitation lies in the computational demands of combining
Physics-Informed Neural Networks (PINNs) in this application. As PINNs
improve prediction capability by including physical regulations in their
models, their implementation demands considerable computational power,
especially in their initial phases. Such computational demands might
limit their application in real-time, for example, in low-resource
settings or in extensive systems. The second crucial point to analyze is
scalability. As encouraging application in residential and small
building power systems looks promising, considerable barriers lie in
attempting to scale up such technology to larger applications, such as
citywide smart power systems. Larger systems have greater variability in
power delivery behavior, load patterns, and time-dependent interactions,
and hence, greater data heterogeneity and system complexity. Such
limitations might impair responsiveness of the model and sacrifice
accuracy in realistic application. Moreover, data-centric approach
relying on extensive collection of quality data using smart meters, IoT,
and renewable power systems is an extra challenge. The model needs to
have ongoing access to consistent, high-resolution data to effectively
train and make necessary real-time interventions. Any inconsistencies,
incomplete data, or poor data quality might compromise prediction
ability and system integrity and thus limit application in places where
smart power monitoring facilities are in their developmental phases.

\subsection{\textbf{6.2. Future Work}}

Addressing the limitations in the envisioned PINNs-DT approach offers
various future directions and improvement options. An important future
avenue is optimizing computational cost. Future studies may focus on
optimizing the training protocol using strategies such as pruning,
parallel computation, and harnessing GPU or TPU capabilities. Such
modifications may result in minimizing computational needs, thus
expanding the model\textquotesingle s viability for real-time and large
implementations. Broadening the approach\textquotesingle s paradigm to
include smart grids on city or countrywide scales is another crucial
future direction. Attaining such an objective means designing modular or
hierarchical structures capable of accommodating greater variability and
heterogeneity in large power systems. Moreover, incorporation of power
generation by multiple power sources and monitoring of greater
variability in dynamic interactions is crucial for efficient scaling of
the envisioned methodology.

\textbf{Ethical Approval}: Not applicable.\\
\textbf{Consent to Participate}: Not applicable.\\
\textbf{Consent to Publish}: Not applicable.\\
\textbf{Funding}: The funding sources were not involved in any aspect of the study or manuscript preparation.\\
\textbf{Competing Interests}: We declare no conflict of interest.\\
\textbf{Author contribution}: All authors have contributed equally to this work.\\
\textbf{Dataset}: Data is available and can be provided over the emails by querying directly to the author at the corresponding author.

\section{References}
\begin{enumerate}
\def\labelenumi{\arabic{enumi}.}
\item
  Hasan, A., \& Aghaei, M. (2025). Digital twin technology: fundamental
  aspects and advances. In Digital Twin Technology for the Energy Sector
  (pp. 25-45). Elsevier.
\item
  Shafto, M., et al. (2010). Modeling, Simulation, Information
  Technology \& Processing Roadmap. NASA.
\item
  Tao, F., \& Zhang, M. (2017). Digital Twin and its Applications in
  Intelligent Manufacturing. Engineering, 5(4), 580-590.
\item
  Varmaghani, A., Matin Nazar, A., Ahmadi, M., Sharifi, A., Jafarzadeh
  Ghoushchi, S., \& Pourasad, Y. (2021). DMTC: Optimize Energy
  Consumption in Dynamic Wireless Sensor Network Based on Fog Computing
  and Fuzzy Multiple Attribute Decision‐Making. Wireless Communications
  and Mobile Computing, 2021(1), 9953416.
\item
  Brosinsky, C., Westermann, D., \& Krebs, R. (2018, June). Recent and
  prospective developments in power system control centers: Adapting the
  digital twin technology for application in power system control
  centers. In 2018 IEEE international energy conference (ENERGYCON) (pp.
  1-6). IEEE.
\item
  Ahmadi, M., Soofiabadi, M., Nikpour, M., Naderi, H., Abdullah, L., \&
  Arandian, B. (2022). Developing a deep neural network with fuzzy
  wavelets and integrating an inline PSO to predict energy consumption
  patterns in urban buildings. Mathematics, 10(8), 1270.
\item
  Sharifi, A., Naeini, H. K., Ahmadi, M., Asadi, S., \& Varmaghani, A.
  (2025). Multi-Objective Optimization of Water Resource Allocation for
  Groundwater Recharge and Surface Runoff Management in Watershed
  Systems. arXiv preprint arXiv:2502.15953.
\item
  Sivalingam, K., et al. (2018). Applications of Digital Twin Technology
  in Renewable Energy Systems. Renewable Energy, 122, 663-673.
\item
  Batty, M. (2018). Digital Twins and Smart Cities. Environment and
  Planning B: Urban Analytics and City Science, 45(3), 349-355.
\item
  Qi, Q., et al. (2018). Digital Twin for Smart Manufacturing: A Review.
  Advanced Manufacturing, 1(1), 1-11.
\item
  Haag, S., \& Anderl, R. (2018). Digital Twin---Proof of Concept. CIRP
  Journal of Manufacturing Science and Technology, 23, 144-152.
\item
  Rojek, I., Mikołajewski, D., Galas, K., \& Piszcz, A. (2025). Advanced
  Deep Learning Algorithms for Energy Optimization of Smart
  Cities.~Energies,~18(2), 407.
\item
  Onile, A. E., Petlenkov, E., Levron, Y., \& Belikov, J. (2024).
  Smartgrid-based hybrid digital twins framework for demand side
  recommendation service provision in distributed power systems. Future
  Generation Computer Systems, 156, 142-156.
\item
  Es-haghi, M. S., Anitescu, C., \& Rabczuk, T. (2024). Methods for
  enabling real-time analysis in digital twins: A literature review.
  Computers \& Structures, 297, 107342.
\item
  Wang, D., Song, Y., Zhang, Y., Jiang, X., Dong, J., Khan, F. N., ...
  \& Zhang, M. (2024). Digital twin of optical networks: a review of
  recent advances and future trends.~Journal of Lightwave Technology.
\item
  Kertha Utama, P., Nashirul Haq, I., Pradipta, J., Putra, A., \&
  Leksono, E. (2024). Microgrid digital twin: implementation of digital
  twin concept based on smart grid architectural model (sgam) and its
  case study.~Irsyad and Pradipta, Justin and Putra, Angga and Leksono,
  Edi, Microgrid digital twin: Implementation of Digital Twin Concept
  Based on Smart Grid Architectural Model (Sgam) and its Case Study.
\item
  Stadtmann, F., Rasheed, A., Kvamsdal, T., Johannessen, K. A., San, O.,
  Kölle, K., ... \& Skogås, J. O. (2023). Digital twins in wind energy:
  Emerging technologies and industry-informed future directions.~IEEE
  Access,~11, 110762-110795.
\item
  Zeng, Y., Hussein, Z. A., Chyad, M. H., Farhadi, A., Yu, J., \&
  Rahbarimagham, H. (2025). Integrating type-2 fuzzy logic controllers
  with digital twin and neural networks for advanced hydropower system
  management.~Scientific Reports,~15(1), 5140.
\item
  Xu, G., \& Guo, T. (2025). Advances in AI-powered civil engineering
  throughout the entire lifecycle.~Advances in Structural Engineering,
  13694332241307721.
\item
  Ahmadi, M., Biswas, D., Paul, R., Lin, M., Tang, Y., Cheema, T. S.,
  ... \& Vrionis, F. D. (2025). Integrating Finite Element Analysis and
  Physics-Informed Neural Networks for Biomechanical Modeling of the
  Human Lumbar Spine.~\emph{North American Spine Society Journal
  (NASSJ)}, 100598.
\item
  Parizad, A., Baghaee, H. R., \& Rahman, S. (2025). Overview of Smart
  Cyber‐Physical Power Systems: Fundamentals, Challenges, and
  Solutions.~Smart Cyber‐Physical Power Systems: Fundamental Concepts,
  Challenges, and Solutions,~1, 1-69.
\item
  Mulpuri, S. K., Sah, B., \& Kumar, P. (2025). Intelligent Battery
  Management System (BMS) with End-Edge-Cloud Connectivity-A
  Perspective.~Sustainable Energy \& Fuels.
\item
  Chen, C., Zhang, L., Zhou, C., \& Luo, Y. (2025). Physics-informed
  explainable encoder-decoder deep learning for predictive estimation of
  building carbon emissions.~Renewable and Sustainable Energy
  Reviews,~213, 115478.
\item
  Chen, D., Lin, X., \& Qiao, Y. (2025). Perspectives for artificial
  intelligence in sustainable energy systems.~Energy, 134711.
\item
  Mittal, P., \& Malik, R. (2025). Optimized Physics-Informed Neural
  Network Framework for Wild Animal Activity Detection and
  Classification with Real Time Alert Message Generation.~International
  Journal on Computational Modelling Applications,~2(1), 42-52.
\item
  Pandiyan, S. V., Gros, S., \& Rajasekharan, J. (2025). Physics
  informed neural network based multi-zone electric water heater
  modeling for demand response.~Applied Energy,~380, 125037.
\item
  Feng, R., Wajid, K., Faheem, M., Wang, J., Subhan, F. E., \& Bhutta,
  M. S. (2025). Uniform Physics Informed Neural Network Framework for
  Microgrid and its application in voltage stability analysis.~IEEE
  Access.
\item
  Habib, A., \& Yildirim, U. (2025). Developing block-based
  physics-informed multi-layered neural network model for simulating the
  inelastic response of base-isolated structures.~Neural Computing and
  Applications, 1-24.
\item
  Nadal, I. V., Nellikkath, R., \& Chatzivasileiadis, S. (2025).
  Physics-Informed Neural Networks in Power System Dynamics: Improving
  Simulation Accuracy.~arXiv preprint arXiv:2501.17621.
\item
  30- Ventura Nadal, I., Nellikkath, R., \& Chatzivasileiadis, S.
  (2025). Physics-Informed Neural Networks in Power System Dynamics:
  Improving Simulation Accuracy.~arXiv e-prints, arXiv-2501.
\item
  Ko, T., Kim, D., Park, J., \& Lee, S. H. (2025). Physics-informed
  neural network for long-term prognostics of proton exchange membrane
  fuel cells.~Applied Energy,~382, 125318.
\item
  Qin, Y., Liu, H., Wang, Y., \& Mao, Y. (2024). Inverse
  physics--informed neural networks for digital twin--based bearing
  fault diagnosis under imbalanced samples.~\emph{Knowledge-Based
  Systems},~\emph{292}, 111641.
\item
  Far, A. Z., Far, M. Z., Gharibzadeh, S., Zangeneh, S., Amini, L., \&
  Rahimi, M. (2024). Artificial intelligence for secured information
  systems in smart cities: Collaborative iot computing with deep
  reinforcement learning and blockchain. arXiv preprint
  arXiv:2409.16444.
\item
  Songhorabadi, M., Rahimi, M., MoghadamFarid, A., \& Kashani, M. H.
  (2023). Fog computing approaches in IoT-enabled smart cities. Journal
  of Network and Computer Applications, 211, 103557.
\item
  Songhorabadi, M., Rahimi, M., Farid, A. M. M., \& Kashani, M. H.
  (2020). Fog computing approaches in smart cities: a state-of-the-art
  review. arXiv preprint arXiv:2011.14732.
\item
  Minerva, R., Lee, G. M., \& Crespi, N. (2020). Digital twin in the IoT
  context: A survey on technical features, scenarios, and architectural
  models.~Proceedings of the IEEE,~108(10), 1785-1824.
\item
  Jafari, M., Kavousi-Fard, A., Chen, T., \& Karimi, M. (2023). A review
  on digital twin technology in smart grid, transportation system and
  smart city: Challenges and future.~IEEE Access,~11, 17471-17484.
\item
  Hegazy Abdelghany Mohamed Ammar, M. (2021). Advanced digital twins for
  conditions monitoring, examinations, diagnosis and predictive
  remaining lifecycles based Artificial Intelligence (Doctoral
  dissertation, Brunel University London).
\item
  Asadi, S., \& Abedini, A. (2016). Numerical study of cooling buildings
  with cylindrical roof using surface-to-airheat exchanger. Journal of
  Mechanical Engineering and Vibration, 7(1), 27-34.
\item
  McIntosh, B. S., Ascough II, J. C., Twery, M., Chew, J., Elmahdi, A.,
  Haase, D., ... \& Voinov, A. (2011). Environmental decision support
  systems (EDSS) development--Challenges and best
  practices.~Environmental Modelling \& Software,~26(12), 1389-1402.
\item
  Tao, F., Zhang, H., Liu, A., \& Nee, A. Y. (2018). Digital twin in
  industry: State-of-the-art.~IEEE Transactions on industrial
  informatics,~15(4), 2405-2415.
\item
  Qian, C., Liu, X., Ripley, C., Qian, M., Liang, F., \& Yu, W. (2022).
  Digital twin---Cyber replica of physical things: Architecture,
  applications and future research directions.~Future Internet,~14(2),
  64.
\item
  Reihanifar, M., Takallou, A., Taheri, M., Lonbar, A. G., Ahmadi, M.,
  \& Sharifi, A. (2024). Nanotechnology advancements in groundwater
  remediation: A comprehensive analysis of current research and future
  prospects.~\emph{Groundwater for Sustainable Development},~\emph{27},
  101330.
\item
  Attari, M. Y. N., Ala, A., Ahmadi, M., \& Jami, E. N. (2024). A highly
  effective optimization approach for managing reverse warehouse system
  capacity across diverse scenarios.~\emph{Process Integration and
  Optimization for Sustainability},~\emph{8}(2), 455-471.
\item
  Asadi, S., Mohammadagha, M., \& Naeini, H. K. (2025). Comprehensive
  Review of Analytical and Numerical Approaches in Earth-to-Air Heat
  Exchangers and Exergoeconomic Evaluations.~\emph{arXiv preprint
  arXiv:2502.08553}.
\item
  Moghim, S., \& Takallou, A. (2023). An integrated assessment of
  extreme hydrometeorological events in Bangladesh.~\emph{Stochastic
  Environmental Research and Risk Assessment},~\emph{37}(7), 2541-2561.
\item
  Asgari, F., Ghoreishi, S. G. A., Khajavi, M., Foozoni, A., Ala, A., \&
  Lonbar, A. G. (2024). Data Analysis of Decision Support for
  Sustainable Welfare in The Presence of GDP Threshold Effects: A Case
  Study of Interactive Data Exploration.~\emph{arXiv preprint
  arXiv:2407.09711}.
\item
  Reihanifar, M., Danandeh Mehr, A., Tur, R., Ahmed, A. T., Abualigah,
  L., \& Dąbrowska, D. (2023). A new multi-objective genetic programming
  model for meteorological drought
  forecasting.~\emph{Water},~\emph{15}(20), 3602.
\item
  Danandeh Mehr, A., Reihanifar, M., Alee, M. M., Vazifehkhah Ghaffari,
  M. A., Safari, M. J. S., \& Mohammadi, B. (2023). VMD-GP: A new
  evolutionary explicit model for meteorological drought prediction at
  ungauged catchments.~\emph{Water},~\emph{15}(15), 2686.
\item
  Hussain, A., Reihanifar, M., Niaz, R., Albalawi, O., Maghrebi, M.,
  Ahmed, A. T., \& Danandeh Mehr, A. (2024). Characterizing
  Inter-Seasonal Meteorological Drought Using Random Effect Logistic
  Regression.~\emph{Sustainability},~\emph{16}(19), 8433.
\item
  Hashmi, R., Liu, H., \& Yavari, A. (2024). Digital twins for enhancing
  efficiency and assuring safety in renewable energy systems: A
  systematic literature review.~Energies,~17(11), 2456.
\item
  Esmaeili, M., Rahimi, M., Pishdast, H., Farahmandazad, D., Khajavi,
  M., \& Saray, H. J. (2024). Machine Learning-Assisted Intrusion
  Detection for Enhancing Internet of Things Security. arXiv preprint
  arXiv:2410.01016.
\item
  Asadi, S., Gharibzadeh, S., Zangeneh, S., Reihanifar, M., Rahimi, M.,
  \& Abdullah, L. (2024). Comparative Analysis of Gradient-Based
  Optimization Techniques Using Multidimensional Surface 3D
  Visualizations and Initial Point Sensitivity. arXiv preprint
  arXiv:2409.04470.
\item
  Cellina, M., Cè, M., Alì, M., Irmici, G., Ibba, S., Caloro, E., ... \&
  Papa, S. (2023). Digital twins: The new frontier for personalized
  medicine?.~Applied Sciences,~13(13), 7940.
\item
  Ahmadi, M., Lonbar, A. G., Naeini, H. K., Beris, A. T., Nouri, M.,
  Javidi, A. S., \& Sharifi, A. (2023). Application of segment anything
  model for civil infrastructure defect assessment.~arXiv preprint
  arXiv:2304.12600.
\item
  Zahid, H., Zulfiqar, A., Adnan, M., Iqbal, M. S., \& Shah, A. (2025).
  Transforming nano grids to smart grid 3.0: AI, digital twins,
  blockchain, and the metaverse revolutionizing the energy ecosystem.
  \emph{Power, Energy and Industry Applications}.
\item
  Sarker, I. H., Janicke, H., Mohsin, A., Gill, A., \& Maglaras, L.
  (2024). Explainable AI for cybersecurity automation, intelligence, and
  trustworthiness in digital twin: Methods, taxonomy, challenges, and
  prospects. \emph{ICT Express}.
  https://doi.org/10.1016/j.icte.2024.05.007
\item
  Idrisov, I. N., Okeke, D., Albaseer, A., Abdallah, M., \& Ibanez, F.
  M. (2025). Leveraging digital twin and machine learning techniques for
  anomaly detection in power electronics dominated grid. \emph{arXiv
  preprint}.
\item
  Meng, X., \& Zhu, J. (2024). Augmenting cybersecurity in smart urban
  energy systems through IoT and blockchain technology within the
  digital twin framework. \emph{Elsevier}.
  https://doi.org/10.1016/j.cse.2024.109613
\item
  Kavousi-Fard, A., Dabholkar, M., Jafari, M., Fardad-Huroubi, M., Deng,
  X., \& Jie, H. (2024). Digital twin for mitigating solar energy
  resources challenges: A perspective. \emph{Solar Energy, 274}, 109645.
  https://doi.org/10.1016/j.solener.2024.109645
\item
  Kabir, M. R., Halder, D., \& Ray, S. (2024). Digital twins for
  IoT-driven energy systems: A survey. \emph{IEEE Access, 12}, 1--14.
\item
  Cali, U., Dimd, B. D., Hajiagilol, P., Moazami, A., Gourisetti, S. N.
  G., \& Lobaccaro, G. (2024). Digital twins: Shaping the future of
  energy systems and smart cities through cybersecurity, efficiency, and
  sustainability. \emph{IEEE Access}.
\item
  Jafari, M., Kavousi-Fard, A., Chen, T., \& Karimi, M. (2024). A review
  on digital twin technology in smart grid, transportation system, and
  smart city: Challenges and future. \emph{IEEE Access, 11}, 1--18.
\item
  Bai, H., \& Wang, Y. (2022). Digital power grid based on digital twin:
  Definition, structure and key technologies.~\emph{Energy
  Reports},~\emph{8}, 390-397.
\item
  Kumari, N., Sharma, A., Tran, B., Chilamkurti, N., \& Alahakoon, D.
  (2023). A comprehensive review of digital twin technology for
  grid-connected microgrid systems: State of the art, potential and
  challenges faced.~\emph{Energies},~\emph{16}(14), 5525.
\item
  Lv, Z., Cheng, C., \& Lv, H. (2023). Digital twins for secure thermal
  energy storage in building.~\emph{Applied Energy},~\emph{338}, 120907.
\item
  Kavousi-Fard, A., Dabbaghjamanesh, M., Jafari, M., Fotuhi-Firuzabad,
  M., Dong, Z. Y., \& Jin, T. (2024). Digital Twin for mitigating solar
  energy resources challenges: A Perspective Review.~\emph{Solar
  Energy},~\emph{274}, 112561.
\item
  Agostinelli, S., Cumo, F., Guidi, G., \& Tomazzoli, C. (2021).
  Cyber-physical systems improving building energy management: Digital
  twin and artificial intelligence.~\emph{Energies},~\emph{14}(8), 2338.
\item
  Nikpour, M., Yousefi, P. B., Jafarzadeh, H., Danesh, K., Shomali, R.,
  Asadi, S., ... \& Ahmadi, M. (2025). Intelligent energy management
  with iot framework in smart cities using intelligent analysis: An
  application of machine learning methods for complex networks and
  systems. Journal of Network and Computer Applications, 235, 104089.
\item
  Liu, S., Tian, J., Ji, Z., Dai, Y., Guo, H., \& Yang, S. (2024).
  Research on multi-digital twin and its application in wind power
  forecasting.~\emph{Energy},~\emph{292}, 130269.
\item
  Asadi, S., \& Abedini, A. (2015). Numerical and Annalytical Review on
  earth to air heat exchanger and Exergoeconomic Analysis. Journal of
  Mechanical Engineering and Vibration, 6(3), 42-46.
\item
  Naseri, F., Gil, S., Barbu, C., Cetkin, E., Yarimca, G., Jensen, A.
  C., ... \& Gomes, C. (2023). Digital twin of electric vehicle battery
  systems: Comprehensive review of the use cases, requirements, and
  platforms.~\emph{Renewable and Sustainable Energy
  Reviews},~\emph{179}, 113280.
\item
  Zhao, K., Liu, Y., Zhou, Y., Ming, W., \& Wu, J. (2025). Digital
  twin-supported battery state estimation based on TCN-LSTM neural
  networks and transfer learning.~\emph{CSEE Journal of Power and Energy
  Systems}.

\item Farhadi Nia, M., Ahmadi, M., \& Irankhah, E. (2025). Transforming dental diagnostics with artificial intelligence: advanced integration of ChatGPT and large language models for patient care. Frontiers in Dental Medicine, 5, 1456208.
 \item Norcéide, F. S., Aoki, E., Tran, V., Nia, M. F., Thompson, C., \& Chandra, K. (2024, December). Positional Tracking of Physical Objects in an Augmented Reality Environment Using Neuromorphic Vision Sensors. In 2024 International Conference on Machine Learning and Applications (ICMLA) (pp. 1031-1036). IEEE.

\item Nia, M. F. (2024). Explore Cross-Codec Quality-Rate Convex Hulls Relation for Adaptive Streaming. arXiv preprint arXiv:2408.09044.

\item Ahmadi, M., Nia, M. F., Asgarian, S., Danesh, K., Irankhah, E., Lonbar, A. G., \& Sharifi, A. (2023). Comparative analysis of segment anything model and u-net for breast tumor detection in ultrasound and mammography images. arXiv preprint arXiv:2306.12510.

\item Jamali, S., \& Abbasalizadeh, M. (2021). Cost‐aware co‐locating of services in Internet of Things by using multicriteria decision making. International Journal of Communication Systems, 34(16), e4962.

\item Abbasalizadeh, M., Vellamchety, K., Rayavaram, P., \& Narain, S. (2024). Dynamic Link Scheduling in Wireless Networks Through Fuzzy-Enhanced Deep Learning. IEEE Open Journal of the Communications Society.
\item Ahmadi, M., Biswas, D., Paul, R., Lin, M., Tang, Y., Cheema, T. S., ... \& Vrionis, F. D. (2025). Integrating finite element analysis and physics-informed neural networks for biomechanical modeling of the human lumbar spine. North American Spine Society Journal (NASSJ), 22, 100598.

\end{enumerate}

\end{document}